\documentclass[10pt,twocolumn,letterpaper]{article}

\usepackage[pagenumbers]{cvpr} % To force page numbers, e.g. for an arXiv version

\usepackage[dvipsnames]{xcolor}
\usepackage{CJKutf8}

\usepackage{times}
\usepackage{epsfig}
\usepackage{graphicx}
\usepackage{amsmath}
\usepackage{amssymb}
\usepackage{mathtools}
\usepackage{multirow}
\usepackage{enumitem}
\usepackage{caption}
\usepackage{subcaption}
\usepackage{array}
\usepackage{colortbl}
\usepackage{booktabs}
\usepackage{bbm}
\usepackage{multicol}
\usepackage{makecell}
\usepackage{xcolor}
\usepackage{xspace}
\usepackage{float}
\usepackage{color}
\usepackage{siunitx}
\usepackage{pifont}
\usepackage{marvosym}

\definecolor{linkColor}{rgb}{0.18,0.39,0.62}
\usepackage{tcolorbox}
\usepackage[pagebackref=true,breaklinks=true,colorlinks,citecolor=linkColor]{hyperref}

\def\confName{CVPR}
\def\confYear{2024}

\title{
How Far Are We to GPT-4V? \\ Closing the Gap to Commercial Multimodal Models with Open-Source Suites
}

\vspace{-1em}
\author{\scalebox{0.91}{
Zhe Chen$^{4,1*\dagger}$, 
Weiyun Wang$^{5,1*\dagger}$, 
Hao Tian$^{2*}$,
Shenglong Ye$^{1*}$, 
Zhangwei Gao$^{1\dagger}$, 
Erfei Cui$^{1\dagger}$, 
Wenwen Tong$^{2}$, }\\
\scalebox{0.91}{
Kongzhi Hu$^{2}$, 
Jiapeng Luo$^{2}$, 
Zheng Ma$^{2}$, 
Ji Ma$^{2}$, 
Jiaqi Wang$^{1}$, 
Xiaoyi Dong$^{6,1}$, 
Hang Yan$^{1}$, 
Hewei Guo$^{2}$, }\\ 
\scalebox{0.91}{
Conghui He$^{1}$,
Botian Shi$^{1}$,
Zhenjiang Jin$^{1}$, 
Chao Xu$^{1}$,
Bin Wang$^{1}$, 
Xingjian Wei$^{1}$, 
Wei Li$^{1}$, 
Wenjian Zhang$^{1}$,} \\
\scalebox{0.91}{
Bo Zhang$^{1}$,
Pinlong Cai$^{1}$,
Licheng Wen$^{1}$,
Xiangchao Yan$^{1}$,
Min Dou$^{1}$,
Lewei Lu$^{2}$,
Xizhou Zhu$^{3,1,2}$,
Tong Lu$^{4}$,
} \\
\vspace{4px}
\scalebox{0.91}{
Dahua Lin$^{6,1}$, 
Yu Qiao$^{1}$, 
Jifeng Dai$^{3,1}$,
Wenhai Wang$^{6,1}$\textsuperscript{\Letter} }\\
\scalebox{0.91}{
$^1$Shanghai AI Laboratory~~~
$^2$SenseTime Research~~~
$^3$Tsinghua University~~~}\\
\vspace{4px}
\scalebox{0.91}{
$^4$Nanjing University~~~
$^5$Fudan University~~~
$^6$The Chinese University of Hong Kong~~~}\\
\small Demo: \url{https://internvl.opengvlab.com} \\
\small Code: \url{https://github.com/OpenGVLab/InternVL} \\
\small Model: \url{https://huggingface.co/OpenGVLab/InternVL-Chat-V1-5}
}

\newcommand{\yes}{\ding{51}}
\newcommand{\no}{\ding{55}}

\newcommand\blfootnote[1]{%
\begingroup
\renewcommand\thefootnote{}\footnote{#1}%
\addtocounter{footnote}{-1}%
\endgroup
}

\definecolor{baselinecolor}{gray}{.9}

\begin{document}
\begin{CJK}{UTF8}{gbsn} 

\def\ie{\emph{i.e.}}
\def\eg{\emph{e.g.}}
\def\etc{\emph{etc}}
\def\etal{{\em et al.~}}
\def\vs{\emph{vs.~}}

\maketitle

\blfootnote{* equal contribution; $\dagger$ interns at OpenGVLab, Shanghai AI Laboratory; \Letter\  corresponding author (wangwenhai@pjlab.org.cn). }

\vspace{-1.2em}
\begin{abstract}

In this report, we introduce InternVL 1.5, an open-source multimodal large language model (MLLM) to bridge the capability gap between open-source and proprietary commercial models in multimodal understanding.
We introduce three simple improvements:
(1) Strong Vision Encoder: we explored a continuous learning strategy for the large-scale vision foundation model---InternViT-6B, boosting its visual understanding capabilities, and making it can be transferred and reused in different LLMs. 
(2) Dynamic High-Resolution: we divide images into tiles ranging from 1 to 40 of 448$\times$448 pixels according to the aspect ratio and resolution of the input images, which supports up to 4K resolution input.
(3) High-Quality Bilingual Dataset: we carefully collected a high-quality bilingual dataset that covers common scenes, document images, 
and annotated them with English and Chinese question-answer pairs, significantly enhancing performance in OCR- and Chinese-related tasks.
We evaluate InternVL 1.5 through a series of benchmarks and comparative studies. 
Compared to both open-source and proprietary commercial models, InternVL 1.5 shows competitive performance, achieving state-of-the-art results in 8 of 18 multimodal benchmarks.

\end{abstract}
\vspace{-0.5em}
\section{Introduction}

\begin{figure}[t!]
    \centering
    \includegraphics[width=1.0\linewidth]{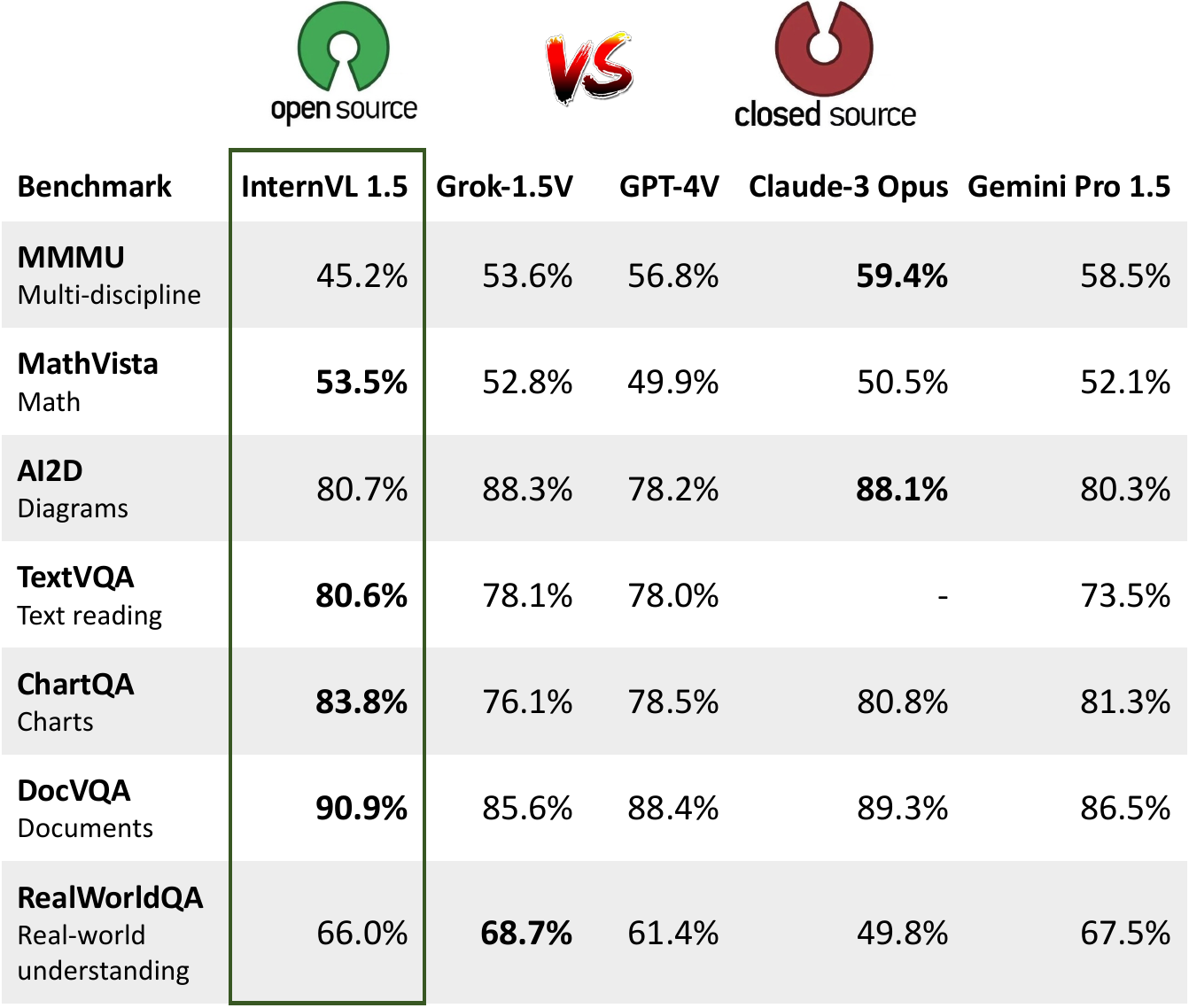}
    \caption{\textbf{InternVL 1.5 versus proprietary commercial models.}
    The results of these benchmarks show that InternVL 1.5 achieves performance comparable to leading proprietary models.
    } 
    \label{fig:teaser}
\end{figure}

Large language models (LLMs) have been instrumental in advancing artificial general intelligence (AGI) systems, demonstrating remarkable abilities in processing open-world language tasks. 
Leveraging the advancements in LLMs, multimodal large language models (MLLMs) \cite{wang2023visionllm, bai2023qwenvl, liu2023llava, liu2023improved, zhu2023minigpt4, chen2023internvl, mckinzie2024mm1,dong2024ixc2,reid2024gemini1_5} have made significant strides, facilitating complex vision-language dialogues and interactions that bridge the gap between textual and visual information. 
Despite these achievements, there remains a noticeable divide between the capabilities of open-source models and proprietary commercial models, \eg, GPT-4V~\cite{gpt4v}, Gemini series~\cite{reid2024gemini1_5, team2023gemini}, and Qwen-VL-Max~\cite{bai2023qwenvl}.

This gap is mainly reflected in the following three aspects:
(1)~\emph{Parameter Scale}: Recent proprietary commercial MLLMs~\cite{gpt4v, reid2024gemini1_5, bai2023qwenvl, step1v2023} typically scales not less than 100 billion parameters, while open-source models commonly employ a 300 million parameter vision foundation model (VFM), which is integrated with either a 7 billion or 13 billion LLMs.
(2)~\emph{Image Resolution}: Proprietary commercial models typically employ a dynamic resolution approach, preserving the original aspect ratio to facilitate detailed scene and document understanding. In contrast, open-source models generally train with fixed resolutions \cite{liu2023improved, chen2023internvl, dong2024ixc2, zhu2023minigpt4, wang2023cogvlm, lu2024deepseekvl}, such as 336$\times$336 and 448$\times$448, leading to a considerable gap in capabilities relative to commercial counterparts.
(3)~\emph{Multilingual Capability}: Proprietary models often leverage extensive multilingual datasets for training, enhancing their performance across diverse languages. However, open-source models predominantly utilize English data, relying on the zero-shot capabilities of LLMs for other languages, \eg~LLaVA-NeXT~\cite{liu2024llavanext}. This results in sub-optimal performance in non-English scene understanding and OCR tasks.

To bridge the gap, we introduce InternVL 1.5, integrating three major improvements to enhance its performance and usability. 
(1) We implement a continuous learning approach to a large-scale VFM---InternViT-6B~\cite{chen2023internvl}, refining it using high-quality image-text data. This process not only enhances the model's ability to understand visual content but also improves its adaptability across various LLMs. 
In addition, using InternLM2-20B~\cite{cai2024internlm2} as the language foundation model also offers robust initial language processing capabilities.
(2) We adopt a dynamic high-resolution strategy that segments images into 448$\times$448 tiles, with the number of tiles ranging from 1 to 40 (\ie, 4K resolution) based on the aspect ratio and resolution of the images. To capture global context, we additionally include a thumbnail view.
(3) We gather a diverse collection of public datasets, covering high-quality natural scenes, charts, documents, and conversations in both English and Chinese. Additionally, we develop a data translation pipeline using open-source LLMs, which can be easily extended to more languages.

These designs endow our model with several advantages:
(1) \emph{Flexible Resolution}: Similar to the ``low'' or ``high'' modes available in GPT-4V~\cite{gpt4v}, InternVL 1.5 enables users to select the optimal resolution for their images, such as using low-resolution for scene subject description and high-resolution (up to 4K resolution) for document understanding, effectively balancing computational efficiency with detail preservation.
(2) \emph{Bilingual Proficiency}: InternVL 1.5 exhibits robust bilingual capabilities, proficiently handling multimodal perception and understanding tasks in both English and Chinese. Notably, in tasks related to Chinese, our model generally outperforms the leading commercial model GPT-4V~\cite{gpt4v}.
(3) \emph{Strong Visual Representation}: By implementing a continuous learning strategy, we enhance the visual representation capabilities of InternViT-6B~\cite{chen2023internvl}, making it robust to flexible input resolution and various visual domains. Benefitting from InternViT-6B's massive parameters, our model achieves a level of visual representation that rivals the linguistic capabilities of LLMs with more than 20 billion parameters. This synergy between visual and linguistic processing endows our system with robust multimodal capabilities.

We evaluated InternVL 1.5 on 18 representative multimodal benchmarks, which are categorized into four specific groups: OCR-related, general multimodal, mathematical, and multi-turn conversation benchmarks.
Compared to both open-source and proprietary models, InternVL 1.5 shows competitive performance, achieving state-of-the-art results in 8 of 18 benchmarks. 
Notably, as shown in Figure~\ref{fig:teaser}, it even surpasses leading proprietary models like Grok-1.5V~\cite{xai2024grokv}, GPT-4V~\cite{gpt4v}, Claude-3 Opus~\cite{claude3series2024}, and Gemini Pro 1.5~\cite{reid2024gemini1_5} in four specific benchmarks, particularly in OCR-related datasets such as TextVQA~\cite{singh2019textvqa}, ChartQA~\cite{masry2022chartqa}, and DocVQA~\cite{mathew2021docvqa}.
This evaluation indicates that InternVL 1.5 has effectively narrowed the gap between open-source models and leading commercial models. We hope that our approach and open-source model weights can contribute to the development of the MLLM community.

\begin{figure}[t!]
    \centering
    \includegraphics[width=0.8\linewidth]{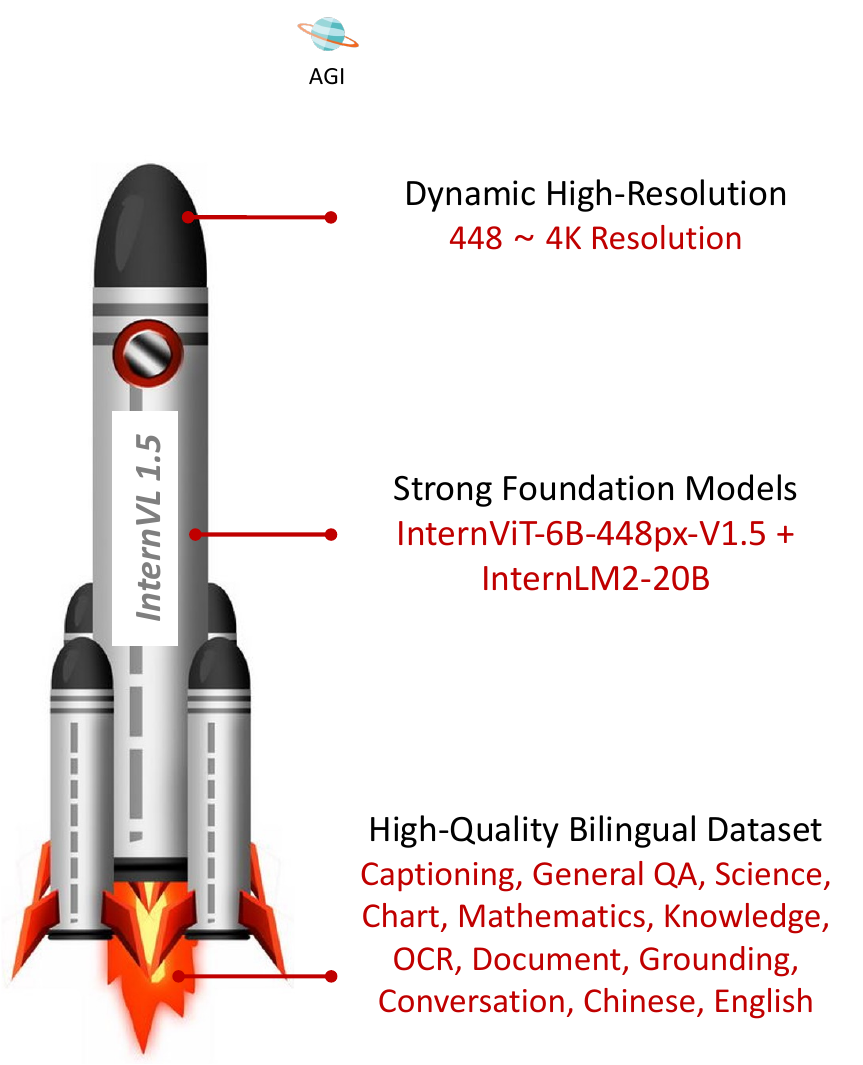}
    \caption{\textbf{Characteristics of InternVL 1.5.}
    InternVL 1.5 features strong visual representation through continuous learning, flexible resolution capabilities, and robust bilingual proficiency in English and Chinese, positioning it as a competitive MLLM.
    } 
    \label{fig:characteristics}
\end{figure}

\section{Related Work}
\label{sec:related_work}

\subsection{Proprietary Commercial MLLMs}
Large language models (LLMs) \cite{openai2023gpt4,touvron2023llama,wei2022chain,touvron2023llama2,zheng2023vicuna,2023internlm,sun2023moss,du2022glm,qwen,baichuan2023baichuan2,taori2023alpaca,wei2023skywork, cai2024internlm2, bi2024deepseekllm} have greatly advanced AGI by enabling complex language tasks previously thought human-exclusive. 
Building on this, the development of proprietary commercial MLLMs represents a significant evolution.
For example, OpenAI’s GPT-4V~\cite{gpt4v} extends GPT-4's capabilities by incorporating visual inputs, allowing it to handle both text and image content, which stands as a significant development in the domain of MLLMs.
Afterward, Google’s Gemini series progresses from Gemini 1.0 \cite{team2023gemini} to Gemini 1.5 \cite{reid2024gemini1_5}, enhancing MLLMs with the ability to process text, images, and audio and support up to 1 million tokens, which boosts performance significantly.
The Qwen-VL-Plus/Max are Alibaba's leading models in the Qwen-VL series \cite{bai2023qwenvl}, renowned for superior capacity in multimodal tasks without needing OCR tools.
Recent advancements in proprietary MLLMs include Anthropic's Claude-3V series~\cite{claude3series2024}, HyperGAI's HPT Pro~\cite{hptpro2024}, Apple’s MM1~\cite{mckinzie2024mm1}, StepFun’s Step-1V~\cite{step1v2023}, and xAI's Grok-1.5V~\cite{xai2024grokv}.

\subsection{Open-Source MLLMs}
The development of open-source MLLMs \cite{zhang2023llama-adapter, zhang2023gpt4roi, wu2023nextgpt, sun2023emu, alayrac2022flamingo, lai2023lisa, li2023otter, li2023monkey, li2023videochat, lin2024moellava, 2023interngpt, liu2023controlllm, tian2024mminterleaved,wang2024allseeingv2, wang2024internvideo2, chen2023videollm} has significantly influenced the AGI landscape by integrating and enhancing capabilities in processing both visual and textual data.
Over the past year, many open-source MLLMs have become well-known, including the LLaVA series~\cite{liu2023llava, liu2023improved, liu2024llavanext}, MiniGPT-4~\cite{zhu2023minigpt4}, VisionLLM~\cite{wang2023visionllm}, Qwen-VL~\cite{bai2023qwenvl}, CogVLM~\cite{wang2023cogvlm}, Shikra~\cite{chen2023shikra}, and others~\cite{wang2023allseeing, peng2023kosmos2, chen2023internvl, dong2024ixc2}.
However, these models are typically trained on images with small, fixed resolutions such as 336$\times$336, or 448$\times$448, which leads to sub-optimal performance on images with unusual aspect ratios or document data.
To address this issue, many approaches have been explored for training on high-resolution images. Currently, there are two common technical routes: one involves designing a dual-branch image encoder~\cite{hong2023cogagent, lv2023kosmos2_5, wei2023vary, luo2024llava_hr,li2024miniGemini}, and the other involves dividing a high-resolution image into many low-resolution tiles~\cite{hu2024mplug_docowl_1_5, li2023otterhd, li2023monkey, lin2023sphinx, liu2024llavanext, liu2024textmonkey, xu2024llava_uhd, ye2023ureader, dong2024xc24khd}.
Despite these explorations in high-resolution training, these open-source models still exhibit significant gaps in understanding documents, charts, and infographics, as well as recognizing scene texts, compared to leading commercial models.

\begin{figure}[t!]
    \centering
    \includegraphics[width=1\linewidth]{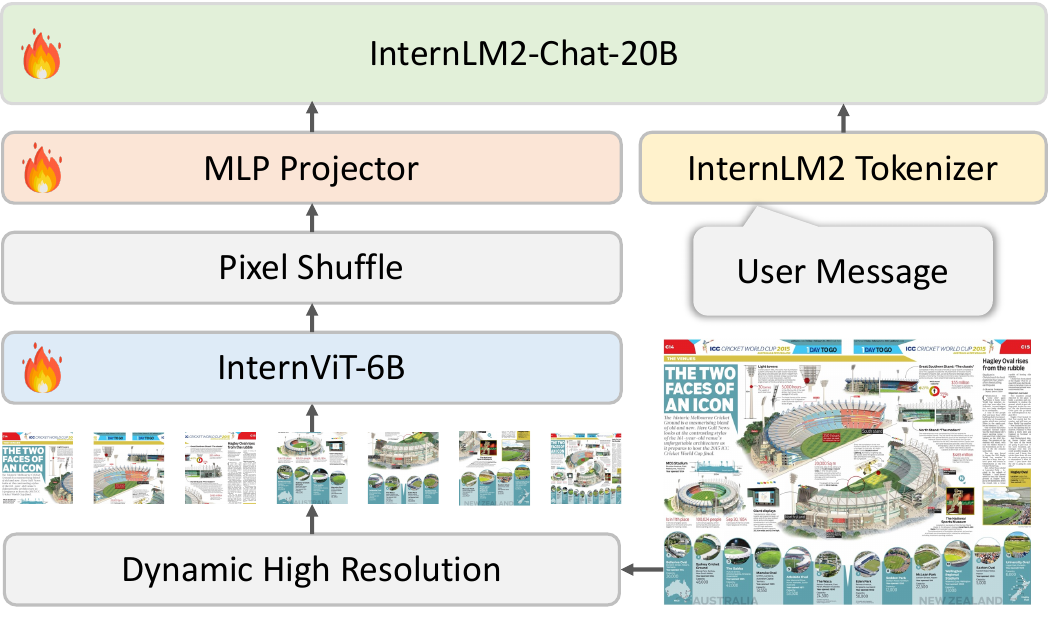}
    \caption{\textbf{Overall Architecture.}
    InternVL 1.5 adopts the ViT-MLP-LLM architecture similar to popular MLLMs~\cite{liu2023improved, liu2024llavanext}, combining a pre-trained InternViT-6B~\cite{chen2023internvl} with InternLM2-20B~\cite{cai2024internlm2} through a MLP projector.
    Here, we employ a simple pixel shuffle to reduce the number of visual tokens to one-quarter.
    } 
    \label{fig:overall}
\end{figure}

\subsection{Vision Foundation Models for MLLMs}

Vision foundation models (VFMs) are a focal point of research within the MLLM community. Currently, models like CLIP-ViT~\cite{radford2021clip} and SigLIP~\cite{zhai2023siglip} are prevalently utilized; however, many studies have been conducted to find the most suitable vision encoders for MLLMs~\cite{lin2023sphinx, tong2024mmvp, luo2024llava_hr, lu2024deepseekvl}. 
For instance, 
Tong \etal\cite{tong2024mmvp} observed notable differences in the visual patterns of CLIP and DINOv2~\cite{oquab2023dinov2}, leading to the development of a mixture-of-features module that combines these two VFMs.
LLaVA-HR~\cite{luo2024llava_hr} introduced a dual-branch vision encoder utilizing CLIP-ViT for low-resolution pathways and CLIP-ConvNext for high-resolution pathways.
Similarly, DeepSeek-VL~\cite{lu2024deepseekvl} adopted a dual vision encoder design, using SigLIP-L for low-resolution images and SAM-B for high-resolution images.
In this report, we propose a continuous learning strategy for our vision foundation model---InternViT-6B~\cite{chen2023internvl}, which continuously boosts the visual understanding capabilities and can be transferred and reused across different LLMs.

\section{InternVL 1.5}
\label{sec:method}

\subsection{Overall Architecture}

As illustrated in Figure~\ref{fig:overall}, InternVL 1.5 employs an architecture akin to widely-used open-source MLLMs, specifically the ``ViT-MLP-LLM'' configuration referenced in various existing studies~\cite{liu2023improved, liu2024llavanext, dong2024ixc2, chen2023internvl, zhu2023minigpt4, liu2023llava, lu2024deepseekvl}. Our implementation of this architecture integrates a pre-trained InternViT-6B~\cite{chen2023internvl} with a pre-trained InternLM2-20B~\cite{cai2024internlm2} using a randomly initialized MLP projector.

During training, we implemented a dynamic resolution strategy, dividing images into tiles of 448$\times$448 pixels in sizes ranging from 1 to 12, based on the aspect ratio and resolution of the input images. 
During testing, this can be zero-shot scaled up to 40 tiles (\ie, 4K resolution).
To enhance scalability for high resolution, we simply employed a pixel shuffle operation to reduce the number of visual tokens to one-quarter of the original.
Therefore, in our model, a 448$\times$448 image is represented by 256 visual tokens.

\subsection{Strong Vision Encoder}

In existing MLLMs~\cite{liu2023improved, liu2023llava, liu2024llavanext, dong2024ixc2, zhu2023minigpt4, bai2023qwenvl,ma2024groma}, the most commonly used vision foundation model is typically a contrastively pre-trained ViT~\cite{openclip,chen2023internvl,radford2021clip,zhai2023siglip}.
However, these ViTs are commonly trained on image-text pairs crawled from the Internet at a fixed low resolution (\eg, 224$\times$224), so their performance degrades when tasked with processing high-resolution images or images from sources other than the Internet, such as document images.

\noindent\textbf{InternViT-6B-448px-V1.2.} To address this issue, 
the InternVL 1.2 update involved continuous pre-training of the InternViT-6B model.
First, we found that the features from the fourth-to-last layer perform best for multimodal tasks, so we directly discarded the weights of the last three layers, reducing InternViT-6B from 48 layers to 45 layers.
Then, we increased the resolution of InternViT-6B from 224 to 448 and integrated it with Nous-Hermes-2-Yi-34B~\cite{young2024yi}.
To equip the model with high-resolution processing and OCR capabilities, both the vision encoder and the MLP were activated for training, utilizing a mix of image captioning~\cite{schuhmann2022laion5b, byeon2022coyo, peng2023kosmos2, chen2015cococaption, singh2019textvqa} and OCR-specific datasets~\cite{gu2022wukong, schuhmann2022laioncoco}.
The newly derived InternViT weights from this process were released as InternViT-6B-448px-V1.2\footnote{\url{https://huggingface.co/OpenGVLab/InternViT-6B-448px-V1-2}}.

\noindent\textbf{InternViT-6B-448px-V1.5.} The development of InternVL 1.5 continues the pre-training of the strong foundation of InternViT-6B-448px-V1.2. In this update, the resolution of training images is expanded from fixed 448$\times$448 to dynamic 448$\times$448, where the basic tile size is 448$\times$448 and the number of tiles ranges from 1 to 12.
Additionally, we enhance the data scale, quality, and diversity of the pre-training dataset, resulting in the powerful robustness, OCR capability, and high-resolution processing capability of our
1.5 version model\footnote{\url{https://huggingface.co/OpenGVLab/InternViT-6B-448px-V1-5}}. Details of the dynamic resolution and training datasets are described in Sections \ref{sec:dhr} and \ref{sec:hbd}.

It is noteworthy that despite the LLM in InternVL 1.5 being changed from Nous-Hermes-2-Yi-34B to InternLM2-20B~\cite{cai2024internlm2}, the InternViT maintained excellent compatibility and portability with the new LLM. 
This suggests that the visual features learned by InternViT-6B during the pre-training stage of MLLMs are broadly applicable and not tightly bound to the specific LLM.

\begin{figure}[t!]
    \centering
    \includegraphics[width=1\linewidth]{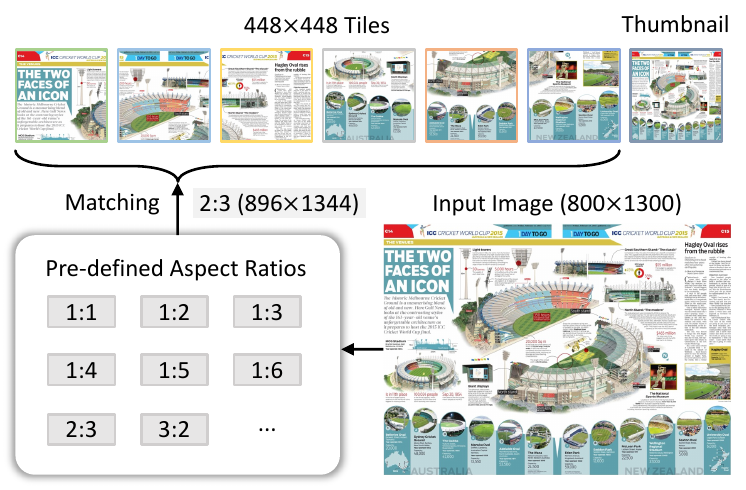}
    \caption{\textbf{Illustration of dynamic high resolution.}
    We dynamically match an optimal aspect ratio from pre-defined ratios, dividing the image into tiles of 448$\times$448 pixels and creating a thumbnail for global context. This method minimizes aspect ratio distortion and accommodates varying resolutions during training.
    } 
    \label{fig:dynamic_high_resolution}
\end{figure}

\subsection{Dynamic High-Resolution}
\label{sec:dhr}
Inspired by UReader~\cite{ye2023ureader}, we adopt a dynamic high-resolution training approach that effectively adapts to the varying resolutions and aspect ratios of input images. This method leverages the flexibility of segmenting images into tiles, enhancing the model's ability to process detailed visual information while accommodating diverse image resolutions.
It mainly consists of the following steps:

\noindent\textbf{Dynamic Aspect Ratio Matching.} As shown in Figure~\ref{fig:dynamic_high_resolution}, to maintain natural aspect ratios during processing, we dynamically match the optimal aspect ratio from a pre-defined set of aspect ratios.
Due to limited computational resources, we allow a maximum of 12 tiles during training. 
Consequently, this set includes all 35 possible combinations of aspect ratios formed by 1 to 12 tiles, such as \{1:1, 1:2, 2:1, 3:1, ..., 2:6\}.
During the matching process, for each input image, we calculate its aspect ratio and compare it with the 35 pre-defined aspect ratios by measuring the absolute difference. 
If multiple pre-defined aspect ratios match (\eg, 1:1 and 2:2), we prioritize the one not exceeding twice the input image's area, thereby preventing excessive enlargement of low-resolution images.

\noindent\textbf{Image Division \& Thumbnail.} Once an appropriate aspect ratio is determined, the image is resized to the corresponding resolution. 
For example, an 800$\times$1300 image will be resized to 896$\times$1344.
The resized image is then divided into tiles of 448$\times$448 pixels.
Alongside the tiles, we include a thumbnail of the entire image to capture the global context. 
This thumbnail is scaled down to 448$\times$448, aiding the model in understanding the overall scene.
Therefore, during training, the number of visual tokens ranges from 256 to 3,328. 
During testing, the number of tiles can increase to a maximum of 40, resulting in 10,496 visual tokens.

\begin{table}[t]\scriptsize
\renewcommand{\arraystretch}{1.2}

\begin{subtable}{0.47\textwidth}
    \centering
    \setlength{\tabcolsep}{0.9mm} 
    \begin{tabular}{l|c|l}
         task & ratio & dataset \\
         \hline
                                      &   & Laion-EN (en)~\cite{schuhmann2022laion5b}, Laion-ZH (zh)~\cite{schuhmann2022laion5b}, COYO (zh)~\cite{byeon2022coyo}, \\ 
         \multirow{-2}{*}{Captioning} & \multirow{-2}{*}{53.9\%}  & GRIT (zh)~\cite{peng2023kosmos2}, COCO (en)~\cite{chen2015cococaption}, TextCaps (en)~\cite{sidorov2020textcaps}   \\
         \rowcolor{gray!15}
                                      &   & Objects365 (en\&zh)~\cite{shao2019objects365}, GRIT (en\&zh)~\cite{peng2023kosmos2},      \\
         \rowcolor{gray!15}
         \multirow{-2}{*}{Detection}  & \multirow{-2}{*}{5.2\%}  & All-Seeing (en\&zh)~\cite{wang2023allseeing} \\
                                      &   & Wukong-OCR (zh)~\cite{gu2022wukong}, LaionCOCO-OCR (en)~\cite{schuhmann2022laioncoco},                      \\
         \multirow{-2}{*}{OCR (large)}& \multirow{-2}{*}{32.0\%}  & Common Crawl PDF (en\&zh)  \\
         \rowcolor{gray!15}
                                      &   & MMC-Inst (en)~\cite{liu2023mmcinst}, LSVT (zh)~\cite{sun2019lsvt}, ST-VQA (en)~\cite{biten2019stvqa}        \\
         \rowcolor{gray!15}
                                      &   & RCTW-17 (zh)~\cite{shi2017rctw17}, ReCTs (zh)~\cite{zhang2019rects}, ArT (en\&zh)~\cite{chng2019art},       \\
         \rowcolor{gray!15}
                                      &   & SynthDoG (en\&zh)~\cite{kim2022synthdog}, COCO-Text (en)~\cite{veit2016cocotext},                           \\
         \rowcolor{gray!15}
                                      &   & ChartQA (en)~\cite{masry2022chartqa}, CTW (zh)~\cite{yuan2019ctw}, DocVQA (en)~\cite{mathew2021docvqa},     \\
         \rowcolor{gray!15}
          \multirow{-5}{*}{OCR (small)}&  \multirow{-5}{*}{8.9\%}  & TextOCR (en)~\cite{singh2021textocr}, PlotQA (en)~\cite{methani2020plotqa}, InfoVQA (en)~\cite{mathew2022infographicvqa}         \\
    \end{tabular}
    \caption{Datasets used in the pre-training stage. 
}
\label{tab:pretraining}
\end{subtable}
\begin{subtable}{0.47\textwidth}
    \setlength\tabcolsep{6.4pt}
    \begin{tabular}{l|l}
task & dataset \\
\hline
Captioning                    & TextCaps (en)~\cite{sidorov2020textcaps}, ShareGPT4V (en\&zh)~\cite{chen2023sharegpt4v}                        \\
\rowcolor{gray!15}
                              & VQAv2 (en)~\cite{goyal2017vqav2}, GQA (en)~\cite{hudson2019gqa}, OKVQA (en)~\cite{marino2019okvqa},            \\
\rowcolor{gray!15}
\multirow{-2}{*}{General QA}  & VSR (en)~\cite{liu2023vsr}, VisualDialog (en)~\cite{das2017visualdialog}                                       \\
\multirow{-1}{*}{Science}     & AI2D (en)~\cite{kembhavi2016ai2d}, ScienceQA (en)~\cite{lu2022scienceqa}, TQA (en)~\cite{kembhavi2017tqa}      \\
\rowcolor{gray!15}
                              & ChartQA (en)~\cite{masry2022chartqa}, MMC-Inst (en)~\cite{liu2023mmcinst}, DVQA (en)~\cite{kafle2018dvqa},     \\
\rowcolor{gray!15}
\multirow{-2}{*}{Chart}       & PlotQA (en)~\cite{methani2020plotqa}, LRV-Instruction (en)~\cite{liu2023lrv-instruction}                       \\

                              & GeoQA+ (en)~\cite{cao2022geoqa_plus}, TabMWP (en)~\cite{lu2022tablemwp}, MathQA (en)~\cite{yu2023mathqa},      \\
\multirow{-2}{*}{Mathematics} & CLEVR-Math/Super (en)~\cite{lindstrom2022clevrmath, li2023superclevr}, Geometry3K (en)~\cite{lu2021geometry3k} \\
\rowcolor{gray!15}
                              & KVQA (en)~\cite{shah2019kvqa}, A-OKVQA (en)~\cite{schwenk2022aokvqa}, ViQuAE (en)~\cite{lerner2022viquae},     \\
\rowcolor{gray!15}
\multirow{-2}{*}{Knowledge}   & Wikipedia (en\&zh)~\cite{he2023wanjuan}                                                                        \\
                              & OCRVQA (en)~\cite{mishra2019ocrvqa}, InfoVQA (en)~\cite{mathew2022infographicvqa}, TextVQA (en)~\cite{singh2019textvqa}, \\ 
                              & ArT (en\&zh)~\cite{chng2019art}, COCO-Text (en)~\cite{veit2016cocotext}, CTW (zh)~\cite{yuan2019ctw},          \\
                              & LSVT (zh)~\cite{sun2019lsvt}, RCTW-17 (zh)~\cite{shi2017rctw17}, ReCTs (zh)~\cite{zhang2019rects},             \\
\multirow{-4}{*}{OCR}         & SynthDoG (en\&zh)~\cite{kim2022synthdog}, ST-VQA (en)~\cite{biten2019stvqa}                                    \\
\rowcolor{gray!15}
Document                      & DocVQA (en)~\cite{clark2017docqa}, Common Crawl PDF (en\&zh)                                                   \\
Grounding                     & RefCOCO/+/g (en)~\cite{yu2016refcoco,mao2016refcocog}, Visual Genome (en)~\cite{krishna2017vg}                            \\
\rowcolor{gray!15}
                              & LLaVA-150K (en\&zh)~\cite{liu2023llava}, LVIS-Instruct4V (en)~\cite{wang2023lvisinstruct4v},                   \\
\rowcolor{gray!15}
                              & ALLaVA (en\&zh)~\cite{chen2024allava}, Laion-GPT4V (en)~\cite{laion_gpt4v_dataset},                            \\
\rowcolor{gray!15}
\multirow{-3}{*}{Conversation}& TextOCR-GPT4V (en)~\cite{textocr_gpt4v_dataset},  SVIT (en\&zh)~\cite{zhao2023svit}                            \\
                              & OpenHermes2.5 (en)~\cite{OpenHermes2_5}, Alpaca-GPT4 (en)~\cite{taori2023alpaca},                              \\
\multirow{-2}{*}{Text-only}   & ShareGPT (en\&zh)~\cite{zheng2023vicuna}, COIG-CQIA (zh)~\cite{bai2024coig}                                    \\
    \end{tabular}
    \centering
        \caption{Datasets used in the fine-tuning stage.
    }
\label{tab:finetuning}
\end{subtable}
\caption{\textbf{Summary of datasets used in InternVL 1.5.} 
To construct large-scale OCR datasets, we utilized PaddleOCR \cite{li2022paddleocr} to perform OCR in Chinese on images from Wukong \cite{gu2022wukong} and in English on images from LAION-COCO \cite{schuhmann2022laioncoco}.
}
\label{tab:dataset}
\end{table}

\subsection{High-Quality Bilingual Dataset}
\label{sec:hbd}

\textbf{Pre-training Dataset.} 
The pre-training dataset utilized in our InternVL 1.5 encompasses a diverse range of publicly accessible sources.
We provide an overview of these datasets in Table~\ref{tab:pretraining}. 
These datasets span multiple tasks, including captioning, which predominantly uses datasets such as Laion-EN \cite{schuhmann2022laion5b}, Laion-ZH \cite{schuhmann2022laion5b}, COYO \cite{byeon2022coyo}, and GRIT \cite{peng2023kosmos2}, constituting 53.9\% of the total data. 
Detection and grounding tasks utilize datasets like Objects365~\cite{shao2019objects365}, GRIT~\cite{peng2023kosmos2}, and All-Seeing~\cite{wang2023allseeing}, making up 5.2\%.
For OCR tasks, we utilized large-scale datasets such as Wukong-OCR, LaionCOCO-OCR, and Common Crawl PDFs, which constitute 32.0\% of our data. 
These datasets were constructed using PaddleOCR \cite{li2022paddleocr} to perform OCR on Chinese images from Wukong \cite{gu2022wukong} and on English images from LaionCOCO \cite{schuhmann2022laioncoco}.
Smaller OCR datasets include MMC-Inst~\cite{liu2023mmcinst}, LSVT~\cite{sun2019lsvt}, ST-VQA~\cite{biten2019stvqa}, RCTW-17~\cite{shi2017rctw17}, ArT~\cite{chng2019art}, and others, accounting for 8.9\% of the data, which focus on more specific or constrained OCR challenges. 
This diverse dataset assembly ensures robust model pre-training of InternVL, catering to varied linguistic and visual elements across tasks.

\begin{figure}[t]\centering
\footnotesize
\begin{minipage}{0.99\columnwidth}\vspace{0mm}    
    \centering
    \begin{tcolorbox}[boxrule=0.2mm]
        \centering
        \hspace{-5mm}
        \begin{tabular}{p{0.99\columnwidth}}
        \hspace{1mm}
        \begin{minipage}{0.99\columnwidth}
        \textbf{System:}

        You are a translator proficient in English and \textcolor{blue}{\{language\}}. Your task is to translate the following English text into \textcolor{blue}{\{language\}}, focusing on a natural and fluent result that avoids ``translationese.'' Please consider these points:

        1.~Keep proper nouns, brands, and geographical names in English.
        
        2.~Retain technical terms or jargon in English, but feel free to explain in \textcolor{blue}{\{language\}} if necessary.
        
        3.~Use \textcolor{blue}{\{language\}} idiomatic expressions for English idioms or proverbs to ensure cultural relevance.
        
        4.~Ensure quotes or direct speech sound natural in \textcolor{blue}{\{language\}}, maintaining the original's tone.
        
        5.~For acronyms, provide the full form in \textcolor{blue}{\{language\}} with the English acronym in parentheses.
        
        \textbf{User:}
        
        Text for translation: \textcolor{blue}{\{text\}}
        
        \textbf{Assistant:}
        
        \textcolor{blue}{\{translation results\}}
        \end{minipage}
        \end{tabular}
    \end{tcolorbox}
    \caption{\textbf{Explanation of our data translation pipeline.} 
    Based on this prompt, we translate English data into Chinese while keeping the language natural and smooth.
    Here, \{language\} represents the target language, \{text\} refers to the original English text, and \{translation results\} indicates the translated text.}
    \label{fig:translation_prompt}
\end{minipage}
\end{figure}

\begin{table*}[t!]
\scriptsize
\centering
\setlength\tabcolsep{0.9pt}
\renewcommand{\arraystretch}{1.15}
\begin{tabular}{l|c|c|ccccc|ccccccccc|c}
                                        & open-   &       & \multicolumn{5}{c|}{OCR-related  Benchmarks} & \multicolumn{9}{c|}{General Multimodal Benchmarks}  & \multicolumn{1}{c}{Math}\\

\multirow{-2}{*}{model} & source & \multirow{-2}{*}{\#param} & DocVQA        & ChartQA       & InfoVQA       & TextVQA       & OCRBench     & MME           & RWQA        & AI2D          & MMMU          & MMB-EN/CN                     & CCB           & MMVet         & SEED           & HallB         & MathVista \\
\hline
GPT-4V~\cite{openai2023gpt4}    &   \no                & $-$ & 88.4          & 78.5	         & $-$           & 78.0          & 645          & 1926.6        &61.4          & 78.2          & 56.8          & 77.0 / 74.4                   & 46.5          & \textbf{67.6} & 71.6          & 46.5          &  49.9     \\ 
Gemini Ultra 1.0~\cite{team2023gemini} & \no           & $-$ & 90.9          & 80.8          & \textbf{80.3} & \textbf{82.3} & $-$          & $-$           &$-$           & 79.5          & \textbf{59.4} & $-$ / $-$                     & $-$           & $-$           & $-$           & $-$           &  53.0     \\
Gemini Pro 1.0~\cite{team2023gemini} & \no             & $-$ & 88.1          & 74.1          & 75.2          & 74.6          & 659          & 1933.4        &$-$           & 73.9          & 47.9          & 73.6 / 74.3                   & 52.5          & 64.3          & 70.7          & 45.2          &  45.2     \\
Gemini Pro 1.5~\cite{reid2024gemini1_5} & \no          & $-$ & 86.5          & 81.3          & 72.7          & 73.5          & $-$          & $-$           &67.5          & 80.3          & 58.5          & $-$ / $-$                     & $-$           & $-$           & $-$           & $-$           &  52.1     \\
Qwen-VL-Max~\cite{bai2023qwenvl} &   \no               & $-$ & \textbf{93.1} & 79.8	         & 73.4          & $-$           & 723          &\textbf{2433.6}&$-$           & 79.3          & 51.3          & 77.6 / 75.7                   & 63.5          & 66.6          & $-$           & 41.2          &  51.0     \\
Qwen-VL-Plus~\cite{bai2023qwenvl} &  \no               & $-$ & 91.4          & 78.1          & $-$           & $-$           & 694          & 2183.4        &$-$           & 75.9          & 45.2          & 67.0 / 70.7                   & 55.1          & 61.1          & 72.7          & 40.6          &  43.3     \\
Claude-3 Opus~\cite{claude3series2024} & \no           & $-$ & 89.3          & 80.8          & $-$           & $-$           & 694          & 1586.8        &49.8          & 88.1          & \textbf{59.4} & 63.3 / 59.2                   & 26.3          & 58.1          & $-$           & 37.8          &  50.5     \\
Claude-3 Sonnet~\cite{claude3series2024} & \no         & $-$ & 89.5          & 81.1          & $-$           & $-$	         & 646          & 1625.9        &51.9          & \textbf{88.7} & 53.1          & 67.8 / 64.2                   & 27.8          & $-$           & $-$           & 41.3          &  47.9     \\
Claude-3 Haiku~\cite{claude3series2024} & \no          & $-$ & 88.8          & 81.7          & $-$           & $-$           & 658          & 1453.2        &$-$           & 86.7          & 50.2          & 60.7 / 57.2                   & 24.5          & $-$           & $-$           & 39.2          &  46.4     \\
HPT Pro~\cite{hptpro2024}  &    \no                    & $-$ & $-$           & $-$           & $-$           & $-$           & $-$          & $-$           &$-$           & $-$           & 52.0          & 77.5 / 76.7                   & $-$           & $-$           & 73.1          & $-$           &  $-$      \\
MM1~\cite{mckinzie2024mm1}     &    \no                & 30B & $-$           & $-$           & $-$           & 73.5          & $-$          & 2069.0        &$-$           & $-$           & 44.7          & 75.1 / $-$~~~~                & $-$           & 48.7          & 72.1          & $-$           &  39.4	   \\
Step-1V~\cite{step1v2023}  &      \no                  & 100B& $-$           & $-$           & $-$           & $-$           & 625          & 2206.4        &$-$           & 79.2          & 49.9          & 80.7 / 79.9                   & \textbf{71.2} & 63.3          & 70.3	       & 48.4          &  44.8     \\
Grok-1.5V~\cite{xai2024grokv}   &   \no                & $-$ & 85.6          & 76.1          & $-$           & 78.1          & $-$          & $-$           &\textbf{68.7} & 88.3          & $-$           & $-$ / $-$                     & $-$           & $-$           & $-$           & $-$           &  52.8     \\
\hline
Text-Monkey~\cite{liu2024textmonkey} & \yes            & 10B & 66.7          & 59.9          & 28.6          & 64.3          & 561          & $-$           &$-$           & $-$           & $-$           & $-$ / $-$                     & $-$           & $-$           & $-$           & $-$           &  $-$      \\
DocOwl-1.5~\cite{hu2024mplug_docowl_1_5}  & \yes       & 8B  & 82.2          & 70.2          & 50.7          & 68.6          & 599          & $-$           &$-$           & $-$           & $-$           & $-$ / $-$                     & $-$           & $-$           & $-$           & $-$           &  $-$      \\
Mini-Gemini~\cite{li2024miniGemini} & \yes             & 35B & $-$           & $-$           & $-$           & ~~74.1*       & $-$          & 2141.0        &$-$           & $-$           & 48.0          & 80.6 / $-$~~~~                & $-$           & 59.3          & $-$           & $-$           &  43.3     \\
LLaVA-NeXT~\cite{liu2024llavanext}  & \yes             & 35B & 84.0          & 68.7          & 51.5          & ~~69.5*       & 574          & 2028.0        &$-$           & 74.9          & 51.1          & 81.1 / 79.0                   & 49.2          & 57.4          & 75.9	       & 34.8          &  46.5     \\
\rowcolor{gray!15}
InternVL 1.2 (ours)         &   \yes                   & 40B & 57.7          & 68.0          & 39.5          & ~~72.5*       & 569          & 2175.4        &67.5          & 79.0          & 51.6          & 82.2 / 81.2                   & 59.2          & 48.9          & 75.6          & 47.6          &  47.7     \\
\rowcolor{gray!15}
InternVL 1.5 (ours)       &     \yes                   & 26B & 90.9          & \textbf{83.8} & 72.5          & 80.6          & \textbf{724} & 2187.8        &66.0          & 80.7          & 45.2          & \textbf{82.2} / \textbf{82.0} & 69.8          & 62.8          & \textbf{76.0} & \textbf{49.3} &  \textbf{53.5} \\

\end{tabular}
\caption{\textbf{Comparison with SoTA models on 16 multimodal benchmarks.}
OCR-related benchmarks include: DocVQA test \cite{mathew2021docvqa}, ChartQA test \cite{masry2022chartqa}, InfographicVQA test \cite{mathew2022infographicvqa}, TextVQA val~\cite{singh2019textvqa}, and OCRBench~\cite{liu2023ocrbench}.
General multimodal benchmarks encompass: MME~\cite{fu2023mme}, RealWorldQA~\cite{xai2024grokv}, AI2D test~\cite{kembhavi2016ai2d}, MMMU val~\cite{yue2023mmmu}, MMBench-EN/CN test~\cite{liu2023mmbench}, CCBench dev~\cite{liu2023mmbench}, MMVet~\cite{yu2023mmvet}, SEED Image~\cite{li2023seed}, and HallusionBench~\cite{guan2023hallusionbench}. 
Additionally, the math dataset includes MathVista testmini~\cite{lu2023mathvista}.
* denotes that Rosetta OCR tokens are used in the testing of TextVQA.
The MME results we report are the sum of the perception and cognition scores.
The results of OCRBench, MMBench, CCBench, and HallusionBench are collected from the OpenCompass leaderboard~\cite{opencompass2023}.
}
\label{tab:sota_results}
\end{table*}

\noindent
\textbf{Fine-tuning Dataset.} 
During the fine-tuning stage, we meticulously selected datasets to enhance model performance across a wide range of multimodal tasks. The datasets used in this phase are summarized in Table~\ref{tab:finetuning}.

For image captioning, we included TextCaps~\cite{sidorov2020textcaps} and bilingual ShareGPT4V~\cite{chen2023sharegpt4v}, which help the model learn to generate descriptive captions in both English and Chinese. In the domain of general QA, datasets such as VQAv2~\cite{goyal2017vqav2}, GQA~\cite{hudson2019gqa}, and VisualDialog~\cite{das2017visualdialog} teach the model to handle diverse question-answering scenarios.

For scientific image understanding, datasets like AI2D \cite{kembhavi2016ai2d}, ScienceQA~\cite{lu2022scienceqa}, and TQA~\cite{kembhavi2017tqa} provide content-rich scenarios to enhance the model's ability to interpret scientific diagrams and texts. Chart interpretation is bolstered by ChartQA~\cite{masry2022chartqa}, MMC-Inst~\cite{liu2023mmcinst}, and PlotQA~\cite{methani2020plotqa}, which train the model to analyze and understand chart images.
Mathematics datasets such as GeoQA+ \cite{cao2022geoqa_plus}, TabMWP \cite{lu2022tablemwp}, and MathQA \cite{yu2023mathqa} introduce complex numerical and geometric problem-solving tasks. Knowledge-based QA benefits from the inclusion of datasets like KVQA~\cite{shah2019kvqa} and bilingual Wikipedia~\cite{he2023wanjuan}, enabling the model to extract and reason with factual information across multiple languages.

For tasks involving OCR, we utilize OCRVQA \cite{mishra2019ocrvqa}, TextVQA \cite{singh2019textvqa}, and several datasets focused on Chinese and English text recognition, such as SynthDoG~\cite{kim2022synthdog}, to improve text recognition from images. Document understanding is advanced through datasets like DocVQA~\cite{mathew2021docvqa} and Common Crawl PDFs, which help the model for real-world document analysis.
Visual grounding is trained using RefCOCO~\cite{yu2016refcoco, mao2016refcocog} and Visual Genome~\cite{krishna2017vg}, aiding the model in precise object localization within images. In the realm of multimodal conversation, datasets like LLaVA-150K~\cite{liu2023llava} and ALLaVA~\cite{chen2024allava} enhance the model's dialogic capabilities by simulating interactive and engaging scenarios.
Lastly, text-only datasets include OpenHermes2.5~\cite{OpenHermes2_5}, Alpaca-GPT4~\cite{taori2023alpaca}, among others~\cite{bai2024coig, zheng2023vicuna}, which are used to maintain the original linguistic capabilities of the LLM.

In summary, these datasets together establish a rich and diverse foundation for fine-tuning, which enhances our model’s ability to handle a wide range of multimodal tasks and ensures its readiness for practical applications.

\noindent\textbf{Data Translation Pipeline.}
As shown in Figure~\ref{fig:translation_prompt}, to enhance our model's multilingual capabilities, we implemented a data translation pipeline. This pipeline utilizes state-of-the-art open-source LLMs~\cite{cai2024internlm2, qwen, young2024yi} or GPT-3.5 to convert English datasets to another language (\eg, Chinese), maintaining consistency and precision in bilingual labeling. Moreover, it can readily expand to encompass more languages by adjusting the language prompt, without relying on manual annotation processes.

In Table~\ref{tab:dataset}, we have annotated the language for each dataset. For a dataset that was originally in English, an annotation as ``zh'' indicates that we have translated it into Chinese using the translation pipeline.
For example, COYO~\cite{byeon2022coyo} and GRIT~\cite{peng2023kosmos2} were originally English datasets, and we have translated them into Chinese.
By leveraging this translation pipeline, the Chinese capabilities of InternVL 1.5 have been greatly enhanced.

\begin{table*}[t!]
\scriptsize
\centering

\renewcommand{\arraystretch}{1.15}
\begin{subtable}{0.685\linewidth}
    \setlength\tabcolsep{1.2pt}
    \begin{tabular}{l|c|c|cccccc|cccccc}
                                              & open-   &       & \multicolumn{6}{c|}{ConvBench (Pairwise Grading)} & \multicolumn{6}{c}{ConvBench (Direct Grading)}  \\
    
    \multirow{-2}{*}{model}             &    source          & \multirow{-2}{*}{\#param} & $R_1$ & $R_2$ & $S_1$ & $S_2$ & $S_3$ & $S_O$   & $R_1$ & $R_2$ & $S_1$ & $S_2$ & $S_3$ & $S_O$   \\
    \hline
    GPT-4V~\cite{openai2023gpt4}           & \no   & $-$ & \textbf{39.51} & \textbf{38.47} & 38.47 & \textbf{39.34} & \textbf{37.61} & \textbf{40.55} & \textbf{7.09} & \textbf{7.30} & \textbf{7.30} & \textbf{7.48} & \textbf{7.12} & \textbf{6.88}      \\ 
    Claude-3 Opus~\cite{claude3series2024} & \no   & $-$ & 36.60 & 37.49 & \textbf{38.99} & 39.17 & 34.32 & 35.70 & 6.54 & 6.75 & 6.53 & 7.04 & 6.68 & 6.32 \\
    Reka Flash~\cite{ormazabal2024reka}    & \no   & $-$ & 25.60 & 24.67 & 25.13 & 27.56 & 21.32 & 26.52 & 6.78 & 6.86 & 6.93 & 7.25 & 6.41 & 6.70  \\ 
    Gemini Pro 1.0~\cite{team2023gemini}   & \no   & $-$ & 8.44  & 8.55  & 9.01  & 9.36  & 7.28  & 8.32  & 4.42 & 4.60 & 5.18 & 4.95 & 3.66 & 4.24  \\
    \hline
    ShareGPT4V-13B~\cite{chen2023sharegpt4v}&\yes  & 13B & 17.56 & 17.45 & 17.85 & 18.72 & 15.77 & 17.68 & 4.85 & 5.03 & 5.16 & 5.06 & 4.86 & 4.67 \\
    LLaVA-1.5-13B~\cite{liu2023improved}   & \yes  & 13B & 16.93 & 18.08 & 20.45 & 18.02 & 15.77 & 15.77 & 4.94 & 5.14 & 5.03 & 5.41 & 4.99 & 4.74 \\
    XComposer2~\cite{dong2024ixc2}         & \yes  & 8B  & 15.83 & 16.41 & 17.16 & 19.06 & 13.00 & 15.25 & 5.82 & 5.98 & 5.98 & 6.17 & 5.78 & 5.66 \\
    mPLUG-Owl2~\cite{ye2023mplug2}         & \yes  & 8B  & 14.93 & 15.83 & 17.50 & 17.16 & 12.82 & 14.04 & 5.04 & 5.17 & 4.98 & 5.38 & 5.14 & 4.91 \\
    Qwen-VL-Chat~\cite{bai2023qwenvl}      & \yes  & 10B & 14.33 & 14.62 & 16.29 & 18.37 & 9.19  & 14.04 & 5.54 & 5.65 & 5.96 & 5.78 & 5.22 & 5.43 \\
    MiniGPT-4~\cite{zhu2023minigpt4}       & \yes  & 8B  & 10.95 & 10.80 & 11.61 & 11.27 & 9.53  & 11.09 & 3.85 & 4.04 & 3.99 & 4.40 & 3.73 & 3.66 \\
    LLaMA-A-V2~\cite{gao2023llama-adapterv2}& \yes & 7B  & 9.04  & 9.59  & 8.84  & 10.92 & 9.01  & 8.49  & 4.77 & 4.91 & 4.77 & 5.47 & 4.48 & 4.64 \\
    \rowcolor{gray!15}
    InternVL 1.2 (ours)                    & \yes  & 40B & 21.17 & 22.41 & 24.96 & 21.31 & 20.97 & 19.93 & 5.49 & 5.69 & 5.80 & 5.88 & 5.39 & 5.29 \\
    \rowcolor{gray!15}
    InternVL 1.5 (ours)                    & \yes  & 26B & 17.65 & 20.22 & 26.00 & 17.33 & 17.33 & 15.08 & 5.60 & 5.76 & 6.11 & 5.93 & 5.25 & 5.43      \\
    % \hline
    
    % \rowcolor{gray!15}

    \end{tabular}
\end{subtable}%
\begin{subtable}{0.4\linewidth}
    \setlength\tabcolsep{1pt}
    \begin{tabular}{l|c|c|cc}
                                              & open-   &       & \multicolumn{2}{c}{MMT-Bench}   \\
    
    \multirow{-2}{*}{model}             &    source          & \multirow{-2}{*}{\#param} & Overall & Overall* \\
    \hline
    GPT-4V~\cite{openai2023gpt4}           & \no   & $-$ & 62.0 & 55.5 \\
    Qwen-VL-Plus~\cite{qwen}               & \no   & $-$ & 62.3 & 56.6 \\
    Gemini Pro 1.0~\cite{team2023gemini}   & \no   & $-$ & 61.6 & 55.1 \\ 
    Claude-3 Haiku~\cite{claude3series2024}& \no    & $-$ & 52.2 & 46.4 \\
    \hline
    LLaVA-NeXT~\cite{liu2024llavanext}     & \yes  & 35B & 60.8 & 56.3 \\
    XComposer2~\cite{dong2024ixc2}         & \yes  & 8B  & 55.7 & 50.0 \\
    BLIP-2-XXL~\cite{li2023blip2}          & \yes  & 12B & 54.8 & 49.1 \\
    Yi-VL-34B~\cite{young2024yi}           & \yes  & 35B & 54.2 & 48.6 \\
    Monkey-Chat~\cite{team2023gemini}      & \yes  & 10B & 53.4 & 46.0 \\
    DeepSeek-VL~\cite{lu2024deepseekvl}    & \yes  & 7B  & 53.2 & 46.5 \\
    CogVLM-Chat~\cite{wang2023cogvlm}      & \yes  & 17B & 51.6 & 44.2 \\
    \rowcolor{gray!15}
    InternVL 1.2 (ours)                    & \yes  & 40B & \textbf{63.4} & \textbf{58.2} \\
    \rowcolor{gray!15}
    InternVL 1.5 (ours)                    & \yes  & 26B & 59.0 & 56.2 \\
    % \hline
    
    \end{tabular}
\end{subtable}%

\caption{\textbf{Comparison with SoTA models on ConvBench and MMT-Bench.
}
ConvBench~\cite{liu2024convbench} is a multi-turn conversation evaluation benchmark designed for MLLMs.
The table presents win rates against humans, where $S_1$, $S_2$, and $S_3$ represent the scores for perception, reasoning, and creation, respectively.  
$R_2$ is calculated as $(S_1 + S_2 + S_3)/3$, reflecting the average performance across three turns. $R_1$ is derived from $(R_2 + S_0)/2$, indicating the model’s overall score.
MMT-Bench~\cite{mmtbench} is a comprehensive benchmark designed to assess MLLMs across massive multimodal tasks requiring expert knowledge and deliberate visual recognition, localization, reasoning, and planning.
The overall score is computed across 162 subtasks, excluding visual recognition as denoted by *.
}
\label{tab:convbench_mmtbench}
\end{table*}

\section{Experiments}
\label{sec:experiments}

\subsection{Implementation Details.}

InternVL 1.5 was developed by integrating the InternViT-6B~\cite{chen2023internvl} vision encoder with the InternLM2-20B~\cite{cai2024internlm2} language model, using a dynamic high-resolution strategy. 
In this approach, images are segmented into 448$\times$448 pixel tiles, with the number of tiles ranging up to 12 based on the image's aspect ratio and resolution during training.
In testing phases, the model could handle up to 40 tiles, equivalent to 4K resolution, demonstrating its adaptability to high-resolution inputs in a zero-shot manner. Notably, we built our model based on the chat version of InternLM2-20B rather than the base model.

The training of InternVL 1.5 was divided into two stages. Initially, the pre-training stage focused on training the InternViT-6B vision encoder and the MLP projector to optimize visual feature extraction. Subsequently, the entire model's 26 billion parameters were fine-tuned to enhance multimodal capabilities.
In both two stages of training, we use a context length of 4096 and adopt the same response formatting prompts as LLaVA 1.5~\cite{li2021improved}.
Additionally, the evaluation was mainly supported by VLMEvalKit~\cite{opencompass2023}.

\subsection{Comparison with State-of-the-Art MLLMs}

\subsubsection{Quantitative Results on 18 Benchmarks}   

In this section, we conduct an extensive evaluation across a series of benchmarks to assess our model's multimodal understanding and reasoning capability. The benchmarks employed in our study are categorized into four distinct types:
OCR-related, general multimodal, mathematical, and multi-turn conversation benchmarks. As depicted in Table~\ref{tab:sota_results}, InternVL 1.5 exhibits leading performance across the majority of these benchmarks.

\noindent\textbf{OCR-related Image Understanding}. 
We evaluate the model performance across four key dimensions of OCR: document comprehension (DocVQA \cite{mathew2021docvqa}), chart understanding (ChartQA \cite{masry2022chartqa}), infographic understanding (InfographicVQA \cite{mathew2022infographicvqa}), and scene text interpretation (TextVQA \cite{singh2019textvqa}). 
Additionally, we employ OCRBench~\cite{liu2023ocrbench} to perform a comprehensive evaluation of the model’s overall OCR capabilities.
As shown in Table \ref{tab:sota_results}, our model demonstrated comparable performance to proprietary models on these benchmarks and significantly outperformed the open-source LLaVA-NeXT~\cite{liu2024llavanext} as well as InternVL 1.2, the predecessor of InternVL 1.5. 
Notably, our model achieves state-of-the-art performance on ChartQA and OCRBench, outperforming all competing proprietary models.

\begin{figure*}[t!]
    \centering
    \vspace{-1em}
    \includegraphics[width=0.93\textwidth]{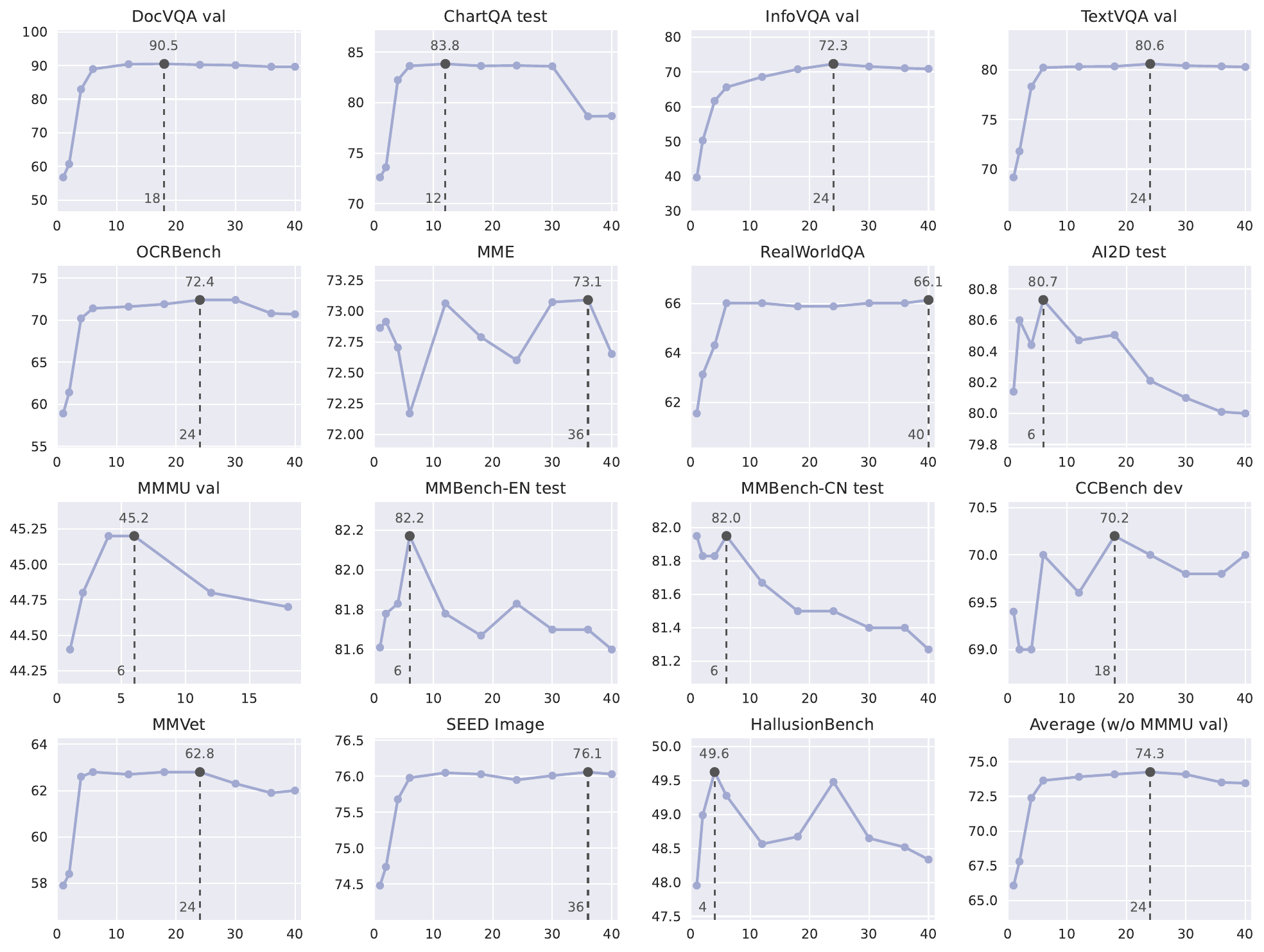}
    \vspace{-0.5em}
\caption{\textbf{Comparison of InternVL 1.5 performance across different image resolutions.}
The X-axis represents the number of tiles, while the Y-axis indicates benchmark performance.
The highest value and its corresponding number of tiles are highlighted.
The scores of MME~\cite{fu2023mme} and OCRBench~\cite{liu2023ocrbench} have been normalized to a maximum score of 100.
We found that although only 1 to 12 tiles were used during training, it is possible to zero-shot scale up to 40 tiles (\ie, 4K resolution) during testing.
Note that since MMMU~\cite{yue2023mmmu} includes multiple images per sample, it may run out of memory when the number of tiles is large. Therefore, we only tested up to 18 tiles maximum, and MMMU was not included when calculating the average score.
    } 
    \label{fig:n_tiles}
\end{figure*}

\noindent\textbf{General Multimodal Evaluation}. 
In addition to OCR-related benchmarks, we tested our model on several general multi-modal benchmarks. 
We used RealWorldQA~\cite{xai2024grokv} to evaluate the model's real-world spatial understanding capabilities.
HallusionBench~\cite{guan2023hallusionbench} was employed to assess its ability to control hallucinations.
Additionally, MMMU \cite{yue2023mmmu} was utilized to evaluate the model's multidisciplinary capabilities, and AI2D~\cite{kembhavi2016ai2d} to assess its understanding of science diagrams. 
We also tested the model's proficiency in Chinese and understanding of Chinese culture with the MMBench-CN test~\cite{liu2023mmbench} and CCBench \cite{liu2023mmbench}, respectively. 
Other comprehensive benchmarks such as MME~\cite{fu2023mme}, MMBench-EN~\cite{liu2023mmbench}, MMVet~\cite{yu2023mmvet}, SEED~\cite{li2023seed}, and MMT-Bench \cite{mmtbench} were also used to assess the model's visual understanding and reasoning abilities.

Compared to other open-source models like Text-Monkey~\cite{liu2024textmonkey}, DocOwl-1.5~\cite{hu2024mplug_docowl_1_5}, and LLaVA-NeXT~\cite{liu2024llavanext}, our InternVL 1.5 significantly closes the gap with proprietary models in these benchmarks. 
Specifically, our model achieves the best performance on HallusionBench~\cite{guan2023hallusionbench}, demonstrating its outstanding ability to reduce hallucinations. 
Moreover, thanks to our high-quality bilingual dataset, our model exhibits robust Chinese language capabilities, significantly surpassing both open-source and proprietary methods on MMBench-CN and CCBench. 
However, while InternVL 1.5 surpasses MM1~\cite{mckinzie2024mm1} and is comparable to Gemini Pro 1.0~\cite{team2023gemini} on MMMU, it shows a slight decline from its predecessor, InternVL 1.2. We attribute this modest decrement to the smaller size of the language model, a phenomenon similarly observed in the MMT-Bench \cite{mmtbench} results, as shown in Table~\ref{tab:convbench_mmtbench}.

\noindent\textbf{Math Reasoning.} 
MathVista~\cite{lu2023mathvista} is a benchmark designed to integrate challenges from various mathematical and visual tasks. Completing these tasks requires a deep understanding of visuals, logical thinking, and math knowledge—areas where many proprietary commercial models encounter significant difficulties. 
As shown in Table \ref{tab:sota_results}, our model outperforms others, including GPT-4V~\cite{gpt4v}, by a clear margin in this benchmark, showcasing its ability to handle mathematically demanding tasks.

\noindent\textbf{Multi-Turn Conversation.}
Compared to single-turn dialogues, multi-turn conversations align more with human preferences. In practical usage, multi-turn dialogue is the preferred mode for general-purpose assistants to engage with humans in solving a variety of tasks. 
Therefore, we opt to utilize ConvBench~\cite{liu2024convbench} for evaluating multi-turn conversations, which progressively assesses the perception, reasoning, and creativity capabilities of MLLMs.
As depicted in Table~\ref{tab:convbench_mmtbench}, InternVL exhibits leading performance among open-source models, albeit still trailing behind GPT-4V by a considerable margin. 
Going forward, we will continue refining InternVL's capabilities in multi-turn conversations.

\begin{figure*}[t]\centering
\begin{minipage}{0.46\linewidth}\vspace{0mm}    
    \centering
    \scriptsize
    \begin{tcolorbox}[boxrule=0.2mm]
        \centering
        \hspace{-5mm}
        \begin{tabular}{p{0.99\columnwidth}}
        \hspace{1mm}
        \begin{minipage}{0.99\columnwidth}
        
        \includegraphics[width=.8\linewidth]{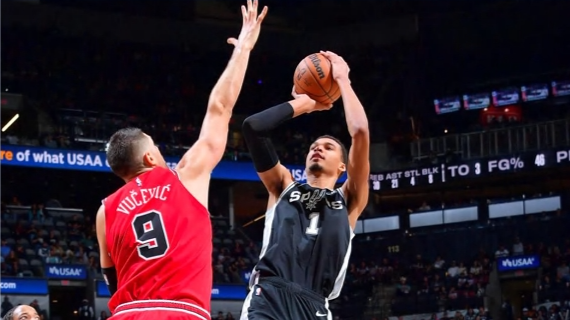}

        \textbf{User:} What action is taking place in the picture? Which color jersey is the attacking player wearing? Explain in detail.
        \vspace{4px}

        \textbf{GPT-4V:} In the image, \textcolor{blue}{we see a basketball game in progress where a player in a black jersey is in the act of shooting the ball. The player is extending their arms upwards towards the basket, and it looks like they're attempting a jump shot.} The player is wearing a black jersey with the word ``SPURS" across the front, which indicates they are a member of the San Antonio Spurs team. There's a defensive player from the opposing team, wearing a red jersey with the number ``9" and the name ``VUCEVIC" on it, who is trying to contest the shot.
        \vspace{4px}
        
        \textbf{InternVL 1.5:} The image depicts a basketball game in progress. \textcolor{blue}{A player in a black jersey, who is likely on the offensive team, is in the act of shooting the basketball. He is airborne, with one arm extended upwards towards the basket, indicating that he is in the process of taking a shot.} The player in the red jersey is on defense, with his arm raised in an attempt to block the shot. The background shows the interior of an arena with spectators and a scoreboard, suggesting that this is a professional or organized basketball match. \textcolor{blue}{The attacking player's jersey is black, and the defending player's jersey is red.}
        
        \end{minipage}
        \end{tabular}
    \end{tcolorbox}
\end{minipage}
\hspace{2em}
\begin{minipage}{0.46\linewidth}\vspace{0mm}    
    \centering
    \scriptsize
    \begin{tcolorbox}[boxrule=0.2mm]
        \centering
        \hspace{-5mm}
        \begin{tabular}{p{0.99\columnwidth}}
        \hspace{1mm}
        \begin{minipage}{0.99\columnwidth}
        
        \includegraphics[width=.46\linewidth]{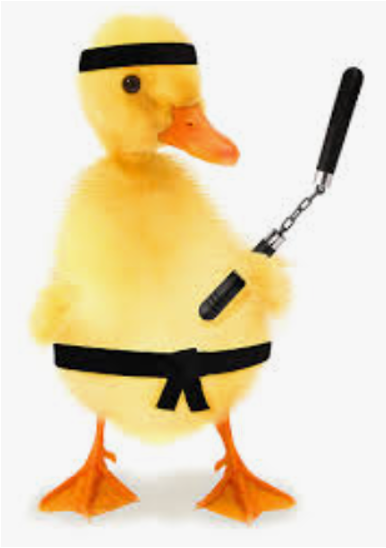} 

        \textbf{User:} Tell me a person who had made a similar pose to the cute duck.
        \vspace{4px}
        
        \textbf{GPT-4V:} \textcolor{red}{I can't identify or make assumptions about a specific person who has made a similar pose to the duck in the image you've uploaded,} but the stance the duck is taking, with what appears to be a martial arts belt and a weapon, is reminiscent of a martial artist demonstrating a pose or ready stance. This is a common pose in many martial arts, as practitioners often pose for photos demonstrating their skills or in the midst of performing kata (a sequence of movements in martial arts). If you're looking for something specific, you might find martial artists or actors who play martial arts roles striking similar poses.
        \vspace{4px}
        
        \textbf{InternVL 1.5:} The pose of the cute duck is reminiscent of a karate stance, which is often associated with martial artists. \textcolor{blue}{One famous person who made a similar pose is Bruce Lee, a martial artist and actor known for his contributions to martial arts and film.}
 
        \end{minipage}
        \end{tabular}
    \end{tcolorbox}
\end{minipage}
\caption{
\textbf{Examples for General QA.}
\textcolor{blue}{Blue} highlights the standout answers, while \textcolor{red}{red} highlights incorrect responses.
}
\label{fig:general_qa}
\end{figure*}

\subsection{Ablation Study}
\noindent\textbf{Larger LLMs need Larger VFMs.}
In this study, we investigate the interplay between LLMs and VFMs. 
The comparison involves two open-source MLLMs, LLaVA-NeXT~\cite{liu2024llavanext} and InternVL 1.2, each equipped with LLMs of 34 billion parameters.
Notably, although both models employ LLMs of the same scale, InternVL 1.2 incorporates a significantly larger VFM, with 6 billion parameters, compared to LLaVA-NeXT's 300 million parameters.
Since the data for LLaVA-NeXT is not available, we created a similar dataset ourselves. Additionally, InternVL 1.2 was trained at a fixed resolution of 448$\times$448, while LLaVA-NeXT used a higher dynamic resolution of $672\times672$. Therefore, this comparison is not entirely fair or equivalent. Nevertheless, the findings still reveal noteworthy insights.
For example, after excluding five OCR-related datasets, ConvBench, and RealWorldQA, InternVL 1.2 outperformed LLaVA-NeXT in 9 out of the remaining 11 datasets. 
This performance difference supports our hypothesis that for a large-scale LLM (\eg, 34B), a larger VFM (\eg, 6B) can effectively improve the model's ability to handle complex multimodal tasks, thereby enhancing the overall performance.

\noindent\textbf{Dynamic Resolution Matters.} 
As shown in Figure \ref{fig:n_tiles}, we investigated the effectiveness of dynamic resolution across various multimodal benchmarks. 
We found that not all tasks require high resolution. Specifically, tasks related to OCR, such as DocVQA, InfoVQA, TextVQA, and OCRBench, benefit from increased resolution. 
However, tasks like AI2D, MMMU, MMBench, and HallusionBench exhibit a slight decline in performance at higher resolutions. 
Overall, InternVL 1.5 demonstrates strong robustness to dynamic resolution. It can adjust the resolution based on the specific requirements of each task, ensuring optimal performance where high resolution is beneficial and conserving resources where it is not.

\subsubsection{Qualitative Results on Different Scenes}
In previous sections, we evaluated our model across various benchmarks and observed its strong performance. In this section, we conduct a qualitative comparison of our model with GPT-4V~\cite{gpt4v} across diverse scenarios, including General QA, OCR-related QA, Scientific Understanding, Chinese Traditional Culture, Object Localization, and Multi-Image Dialogue.
We aim to demonstrate the practicality and versatility of our model in real-world applications, offering insights from the perspective of actual user experience.

\noindent\textbf{General QA}.
To compare the general capabilities of InternVL 1.5 and GPT-4V, we first conducted an experiment involving simple user queries with images requiring general knowledge. 
As shown on the left side of Figure~\ref{fig:general_qa}, both two models respond accurately to the query, showcasing their proficiency in general topics. 
As shown on the right side of Figure~\ref{fig:general_qa}, GPT-4V may excessively refuse to answer some questions due to its involvement in personal privacy.

\begin{figure*}[t]\centering
\begin{minipage}{0.46\linewidth}\vspace{0mm}    
    \centering
    \scriptsize
    \begin{tcolorbox}[boxrule=0.2mm]
        \centering
        \hspace{-5mm}
        \begin{tabular}{p{0.99\columnwidth}}
        \hspace{1mm}
        \begin{minipage}{0.99\columnwidth}
        
        \includegraphics[width=.88\linewidth]{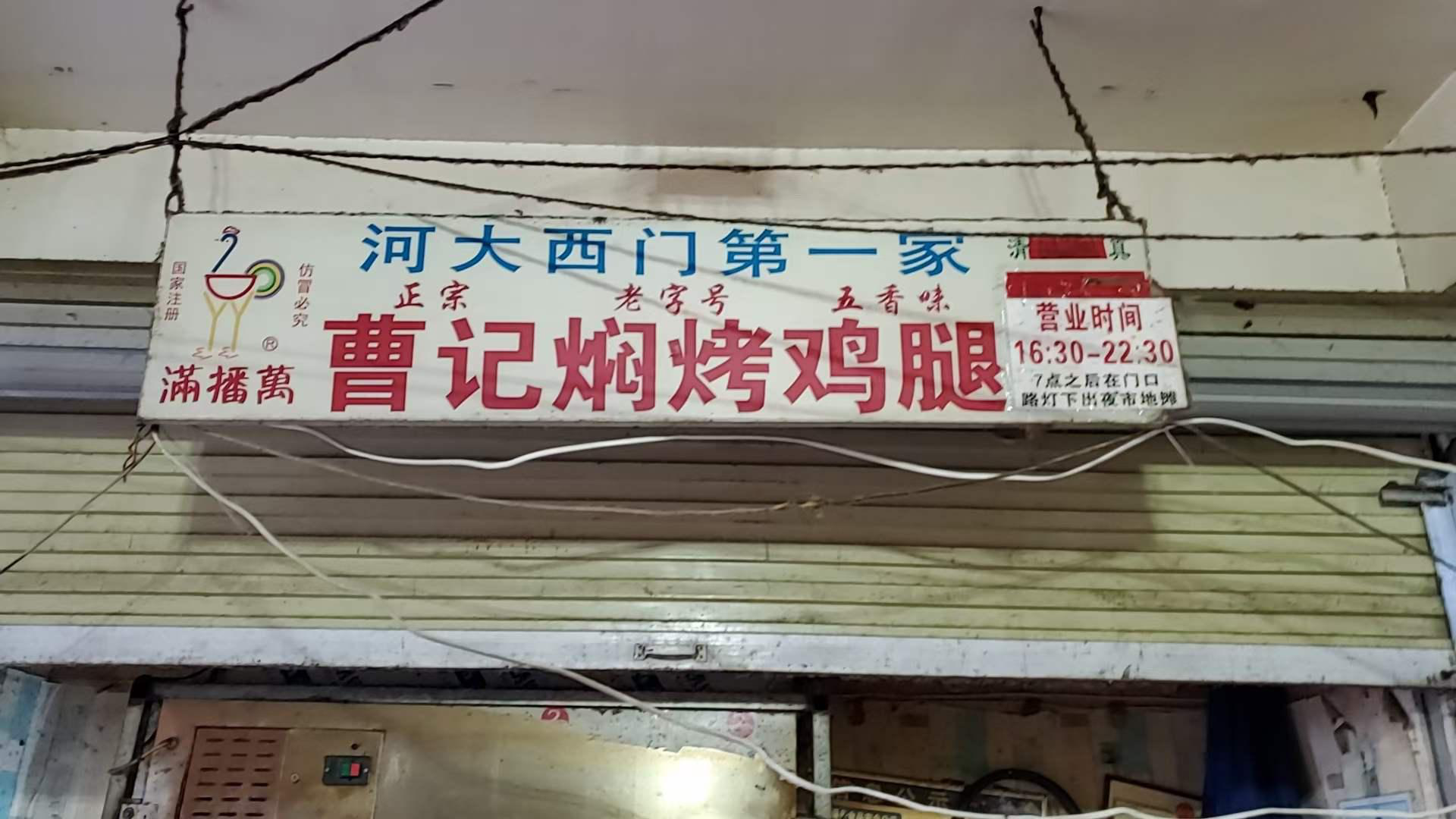} 

        \textbf{User:} Describe this image in detail.
        \vspace{4px}
        
        \textbf{GPT-4V:} The image shows a signboard with Chinese characters on it. From the text, we can deduce that the sign is likely for a business. The sign features a design that includes \textcolor{blue}{a martini glass with a flamingo}, indicating that the establishment might serve drinks. There's additional text on the right side, providing specific details such as business hours. It appears to be a nighttime business, likely a bar or restaurant, \textcolor{blue}{operating from 16:30 to 22:30.} The other text provides further information about the services or specialties of the place. Unfortunately, \textcolor{red}{I can't provide a translation of the text, as my capabilities don't include translating text within images.}
        \vspace{4px}
        
        \textbf{InternVL 1.5:} The image shows a signboard with Chinese text, indicating a business establishment. The signboard reads:

         - \textcolor{blue}{``河大西门第一家"} which translates to ``The first shop at the west gate of the river."
         
         - \textcolor{blue}{``正宗 老字号 五香味"} which means ``Genuine, old brand, five-spice flavor."
         
         - \textcolor{blue}{``曹记焖烤鸡腿"} which translates to ``Cao's roasted and baked chicken legs."
         
         - \textcolor{blue}{``营业时间 16:30-22:30"} indicating the business hours from 4:30 PM to 10:30 PM.
         
         - \textcolor{blue}{``7点之后 在门口 路灯下出夜市地摊"} which suggests that after 7 PM, the shop operates in the doorway under the streetlights as a night market stall.
         
        The signboard also includes \textcolor{blue}{a logo that appears to be a stylized rooster}, which might be the brand or shop's emblem. The overall style of the signboard is quite traditional and straightforward, commonly seen in Chinese street markets or small businesses.
        
        \end{minipage}
        \end{tabular}
    \end{tcolorbox}
\end{minipage}
\hspace{2em}
\begin{minipage}{0.46\linewidth}\vspace{0mm}    
    \centering
    \scriptsize
    \begin{tcolorbox}[boxrule=0.2mm]
        \centering
        \hspace{-5mm}
        \begin{tabular}{p{0.99\columnwidth}}
        \hspace{1mm}
        \begin{minipage}{0.99\columnwidth}
        
        \includegraphics[width=.87\linewidth]{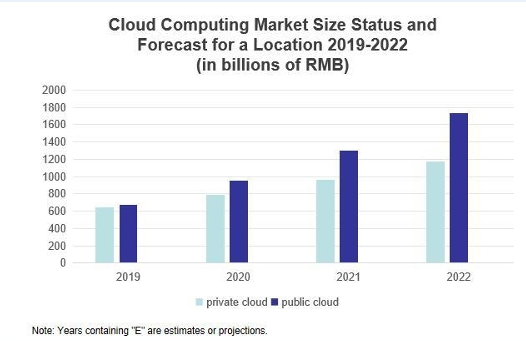} 

        \textbf{User:} What is the difference in total cloud computing market size between the year with the highest revenue and the year with the lowest revenue?
        \vspace{4px}

        \textbf{GPT-4V:} To find the difference in total cloud computing market size between the year with the highest revenue and the year with the lowest revenue, we need to:
        
        1.~Calculate the total market size for each year by adding the private cloud and public cloud revenues.
        
        2.~Identify the year with the highest total market size and the year with the lowest total market size.
        
        3.~Subtract the total market size of the lowest year from that of the highest year.
        
        Looking at the chart, visually, \textcolor{blue}{it seems 2022 has the highest total market size and 2019 has the lowest. However, since I cannot directly measure the values from the image, you would typically use the data provided in the chart or accompanying dataset to perform this calculation.}
        Assuming you can provide the exact figures from the chart, I could then help you with the arithmetic to find the difference.
        \vspace{4px}
        
        \textbf{InternVL 1.5:} Based on the image, the highest revenue is projected for the year 2022 with the \textcolor{blue}{private cloud at 1200 billion RMB} and the \textcolor{blue}{public cloud at 1700 billion RMB}, totaling \textcolor{blue}{2900 billion RMB}. The lowest revenue is shown for the year 2019 with \textcolor{blue}{the private cloud at 650 billion RMB} and \textcolor{blue}{the public cloud at 670 billion RMB}, totaling \textcolor{blue}{1320 billion RMB}. The difference in total cloud computing market size between the year with the highest revenue (2022) and the year with the lowest revenue (2019) is \textcolor{blue}{1580 billion RMB}.
        
        \end{minipage}
        \end{tabular}
    \end{tcolorbox}
\end{minipage}
\caption{
\textbf{Examples for OCR-related QA.}
\textcolor{blue}{Blue} highlights the standout answers, while \textcolor{red}{red} highlights incorrect responses.
}
\vspace{-0.5em}
\label{fig:ocr_qa}
\end{figure*}

\noindent\textbf{OCR-Related QA}. We conducted an evaluation to compare the OCR capabilities of our InternVL 1.5 model against GPT-4V. 
On the left side of Figure~\ref{fig:ocr_qa}, the first prompt aimed to measure the models' ability to understand Chinese scenes. In this instance, GPT-4V cannot extract all useful information in the image.
On the right side of Figure~\ref{fig:ocr_qa}, both GPT-4V and our model have good performance on chart understanding.

\begin{figure*}[t]\centering

\begin{minipage}{0.37\linewidth}\vspace{0mm}    
    \centering
    \scriptsize
    \begin{tcolorbox}[boxrule=0.2mm]
        \centering
        \hspace{-5mm}
        \begin{tabular}{p{0.99\columnwidth}}
        \hspace{1mm}
        \begin{minipage}{0.99\columnwidth}
        
        \includegraphics[width=.99\linewidth]{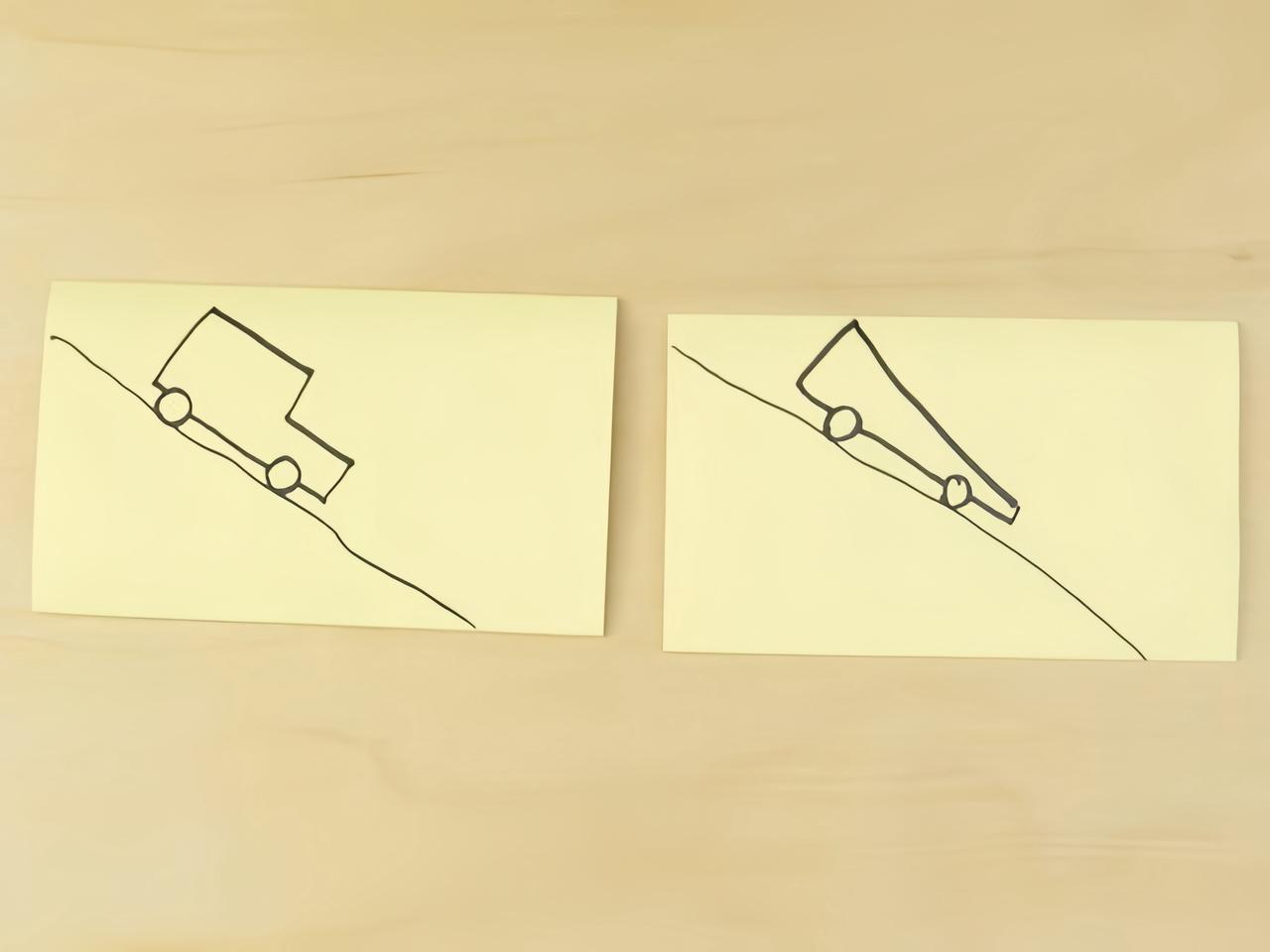} 

        \textbf{User:} Which car in the picture is more aerodynamic?
        \vspace{4px}

        \textbf{GPT-4V:} Based on the image you’ve provided, determining which car is more aerodynamic involves analyzing the shape and design of each vehicle. \textcolor{blue}{The car on the right appears to have a more streamlined shape, with a sloped front that would likely reduce air resistance better than the boxy design of the car on the left.} Therefore, the car on the right would be considered more aerodynamic. 
        \vspace{4px}
        
        \textbf{InternVL 1.5:} The car on the right appears to be more aerodynamic. \textcolor{blue}{This is inferred from the shape of the car, which has a more streamlined and elongated form. The design includes a pointed front end and a tapered rear, which are common features in aerodynamic design to reduce air resistance.} The car on the left, with its boxy shape, does not exhibit the same level of aerodynamic efficiency.
        
        \end{minipage}
        \end{tabular}
    \end{tcolorbox}
\end{minipage}
\hspace{2em}
\begin{minipage}{0.55\linewidth}\vspace{0mm}    
    \centering
    \scriptsize
    \begin{tcolorbox}[boxrule=0.2mm]
        \centering
        \hspace{-5mm}
        \begin{tabular}{p{0.99\columnwidth}}
        \hspace{1mm}
        \begin{minipage}{0.99\columnwidth}
        \includegraphics[width=.5\linewidth]{figure/mathqa_1.PNG}
        
        \textbf{User:} If the pH outside the cell decreases, would you expect the amount of amino acids transported into the cell to increase or decrease?
        \vspace{4px}

        \textbf{GPT-4V:} The image depicts a sodium-glucose symporter, which is a type of secondary active transport mechanism. This transport mechanism is specifically for glucose and does not directly provide information about amino acid transporters.
        However, \textcolor{blue}{speaking generally about amino acid transporters, if the pH outside the cell decreases, the effect on amino acid transport would depend on the type of amino acid transporter involved. Some amino acid transporters are proton-dependent, meaning that they use the proton gradient (which increases as pH decreases) to transport amino acids into the cell. For these transporters, a decrease in extracellular pH could potentially increase the transport of amino acids into the cell.}
        To give a precise answer, one would need specific information about the type of amino acid transporter present in the cell membrane. Since this information is not provided in the image or the description, the impact of a decrease in extracellular pH on amino acid transport cannot be determined from the provided content alone. 
        \vspace{4px}
        
        \textbf{InternVL 1.5:} Based on the image, if the pH outside the cell decreases, we would expect the amount of amino acids transported into the cell to increase. This is because the diagram shows a gradient of amino acids moving from an area of lower concentration (outside the cell) to an area of higher concentration (inside the cell), which is a typical representation of active transport. \textcolor{blue}{Active transport is a process that moves substances against their concentration gradient, often using energy, and it is typically more efficient at higher concentrations of the substance being transported. Therefore, a decrease in pH outside the cell would likely increase the concentration gradient, leading to an increase in the transport of amino acids into the cell.}
        
        \end{minipage}
        \end{tabular}
    \end{tcolorbox}
\end{minipage}
\caption{
\textbf{Examples for Scientific Understanding.}
\textcolor{blue}{Blue} highlights the standout answers.
}
\label{fig:math_qa}
\end{figure*}

\begin{figure*}[t]\centering
\begin{minipage}{0.46\linewidth}\vspace{0mm}    
    \centering
    \scriptsize
    \begin{tcolorbox}[boxrule=0.2mm]
        \centering
        \hspace{-5mm}
        \begin{tabular}{p{0.99\columnwidth}}
        \hspace{1mm}
        \begin{minipage}{0.99\columnwidth}
        
        \includegraphics[width=.82\linewidth]{figure/culture_1.jpg}  

        \textbf{User:} 图中所示的传统表演形式为?
        \vspace{4px}

        \textbf{GPT-4V:} 这张图片展示的是\textcolor{blue}{中国的传统木偶戏，也被称为“提线木偶”。这种表演艺术形式在中国有着悠久的历史，通常以精美的木偶和复杂的操作技巧著称。}木偶戏在中国的各个地区有不同的风格和表演特色。图片中的木偶服饰绚丽，表演者通过线控制木偶的动作，进行故事叙述和表演。 
        \vspace{4px}
        
        \textbf{InternVL 1.5:} 图中所示的传统表演形式为\textcolor{blue}{布袋戏，属于木偶戏的一种，布袋戏是一种古老的汉族民间戏曲表演艺术形式，布袋戏（又称布袋木偶戏、手操傀儡戏、掌中戏、小笼、指花戏）是一种源于17世纪中国福建泉州，流行于台湾、福建、广东等地的传统戏剧艺术。}表演者用双手操纵布袋内的木偶，通过手指的动作来控制木偶的动作和表情，同时配合音乐、唱腔和道白，来讲述故事。在中国以及世界华人地区都有流传。 
        
        \end{minipage}
        \end{tabular}
    \end{tcolorbox}
\end{minipage}
\hspace{2em}
\begin{minipage}{0.46\linewidth}\vspace{0mm}    
    \centering
    \scriptsize
    \begin{tcolorbox}[boxrule=0.2mm]
        \centering
        \hspace{-5mm}
        \begin{tabular}{p{0.99\columnwidth}}
        \hspace{1mm}
        \begin{minipage}{0.99\columnwidth}
        
        \includegraphics[width=.58\linewidth]{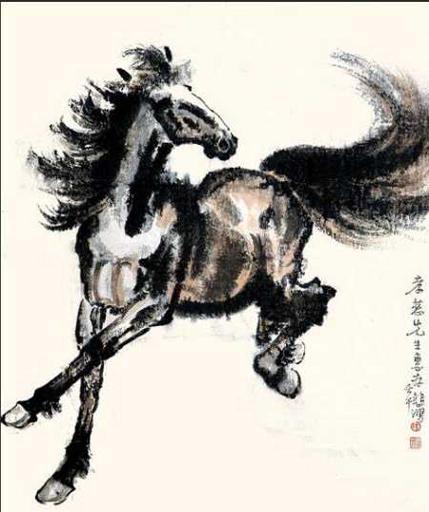} 

        \textbf{User:} 图中是谁的作品? Options: A. 徐悲鸿 B. 齐白石 C. 张大千 D. 刘海粟
        \vspace{4px}

        \textbf{GPT-4V:} 这幅画的风格似乎是徐悲鸿的，他以表现马的高超技巧而闻名。\textcolor{blue}{徐悲鸿的马画特点是线条流畅而有力，体现了马的动态美和生命力。这幅画中马的表现与徐悲鸿的风格相吻合，所以正确答案是 A. 徐悲鸿。} 
        \vspace{4px}
        
        \textbf{InternVL 1.5:} 这幅画看起来像是一幅中国画，特别是马的描绘方式和风格。\textcolor{blue}{徐悲鸿是中国著名的画家，以画马而闻名，他的画作通常具有很强的表现力和动感。从图中马的姿态和表现手法来看，与徐悲鸿的风格非常相似。徐悲鸿的作品通常以大写意的方式呈现，注重表现物象的精神和气质。图中马的神态和动态捕捉得非常生动，体现了徐悲鸿对马匹的深刻理解和精湛的绘画技巧。因此，根据提供的选项，图中作品很可能是徐悲鸿的作品。}
        
        \end{minipage}
        \end{tabular}
    \end{tcolorbox}
\end{minipage}
\caption{\textbf{Examples for Chinese Traditional Culture.} 
\textcolor{blue}{Blue} highlights the standout answers. 
}
\label{fig:chinese}
\end{figure*}

\begin{figure*}[t]\centering
\begin{minipage}{0.46\linewidth}\vspace{0mm}    
    \centering
    \scriptsize
    \begin{tcolorbox}[boxrule=0.2mm]
        \centering
        \hspace{-5mm}
        \begin{tabular}{p{0.99\columnwidth}}
        \hspace{1mm}
        \begin{minipage}{0.99\columnwidth}
        
        \includegraphics[width=0.98\linewidth]{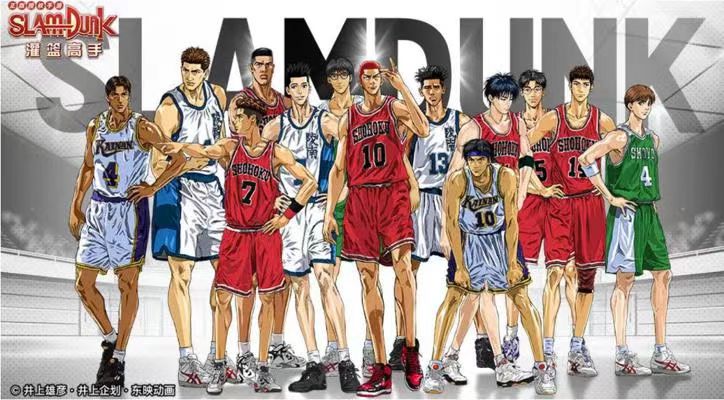}  

        \textbf{User:} Tell me the locations of all the number 10 jerseys in the picture.
        
        \vspace{4px}

        \textbf{GPT-4V:} In the image, \textcolor{blue}{there are two players wearing number 10 jerseys. The first one is located towards the center, slightly to the right while facing the image; he's wearing a red jersey} and \textcolor{red}{is the second person from the right in the group wearing red and white uniforms.} The second player is wearing a light blue jersey and \textcolor{red}{is positioned on the far left side of the image; he's the third person from the left in the group wearing blue and white uniforms.} 
        \vspace{4px}
        
        \textbf{InternVL 1.5:} In the image, there are two number 10 jerseys visible:
        \vspace{1px}
        
        1. \textcolor{blue}{The first player wearing the number 10 jersey is located in the center of the image. He is wearing a red jersey with the number 10 on it}, and he is \textcolor{red}{standing with his hands on his hips.}
        \vspace{1px}
        
        2. \textcolor{blue}{The second player with the number 10 jersey is on the far right of the image. He is wearing a blue jersey with the number 10 on it and is crouching down with his hands on his knees.} 
        
        \end{minipage}
        \end{tabular}
    \end{tcolorbox}
\end{minipage}
\hspace{2em}
\begin{minipage}{0.46\linewidth}\vspace{0mm}    
    \centering
    \scriptsize
    \begin{tcolorbox}[boxrule=0.2mm]
        \centering
        \hspace{-5mm}
        \begin{tabular}{p{0.99\columnwidth}}
        \hspace{1mm}
        \begin{minipage}{0.99\columnwidth}
        
        \includegraphics[width=.51\linewidth]{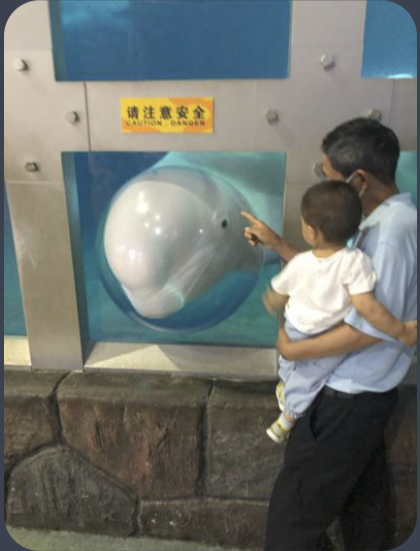} 

        \textbf{User:} What is the man in the picture doing with his right hand?
        \vspace{4px}

        \textbf{GPT-4V:} \textcolor{blue}{The man in the picture is pointing with his right hand towards a beluga whale that appears to be behind a glass barrier. He seems to be doing this to show the whale to the child he is holding, likely to engage the child's interest or to teach the child about the whale.}
        \vspace{4px}
        
        \textbf{InternVL 1.5:} \textcolor{blue}{The man in the picture is pointing at a large dolphin through a glass barrier. He is likely making this gesture to draw attention to the dolphin, either for the child he is holding or for others around them. This could be to educate the child about the dolphin, to highlight a specific behavior or feature of the dolphin, or simply to share the experience with others.} The gesture is a common way to direct someone's attention to something of interest. 
        
        \end{minipage}
        \end{tabular}
    \end{tcolorbox}
\end{minipage}
\caption{\textbf{Examples for Object Localization.} 
\textcolor{blue}{Blue} highlights the standout answers, while \textcolor{red}{red} highlights incorrect responses.
}
\label{fig:localization}
\end{figure*}

\begin{figure*}[t]\centering
\begin{minipage}{0.97\linewidth}\vspace{0mm}    
    \centering
    \scriptsize
    \begin{tcolorbox}[boxrule=0.2mm]
        \centering
        \hspace{-5mm}
        \begin{tabular}{p{0.99\columnwidth}}
        \hspace{1mm}
        \begin{minipage}{0.99\columnwidth}
        
        \includegraphics[width=0.9\linewidth]{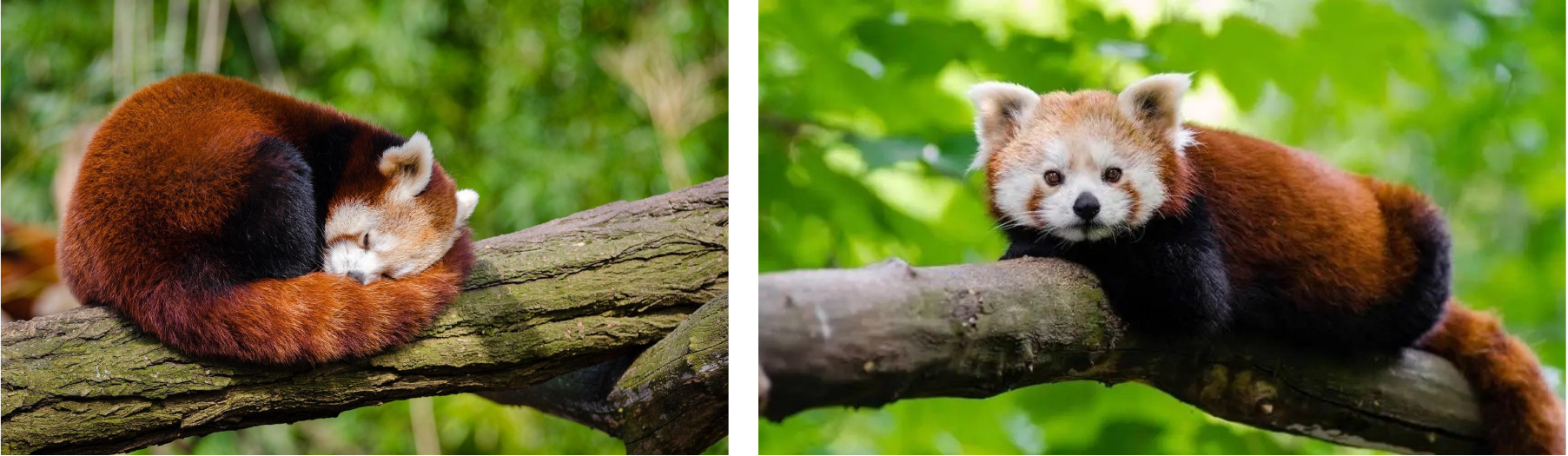}  

        \textbf{User:} What are the similarities and differences between these two pictures?
        
        \vspace{4px}

        \textbf{GPT-4V:} \textcolor{blue}{The similarities between the two images are that they both feature red pandas, a species known for their reddish-brown fur, bushy tails, and adorable faces. In both images, the red pandas are resting on tree branches with green foliage in the background, highlighting their arboreal (tree-dwelling) nature.}
        
        \vspace{2px}
        \textcolor{blue}{As for the differences, the first image shows a red panda in a curled-up position, appearing to be asleep with its tail wrapped around its body for warmth.} This posture also serves as camouflage, making the panda resemble a tree branch or part of the foliage. The angle of the shot is from above the animal, and the focus is on the texture and patterns of the fur.

        \vspace{2px}
        \textcolor{blue}{In contrast, the second image captures the red panda in a more alert and awake state, with its head raised and looking toward the camera.} This posture allows a clear view of the panda's facial features, such as its ears, eyes, and the white markings. The angle of the second image is more direct and at the same level as the animal, which gives a better sense of the panda's overall appearance and stance.
        \vspace{4px}
        
        \textbf{InternVL 1.5:} \textcolor{blue}{The two pictures both feature red pandas, which are small, arboreal mammals native to the temperate forests of the Himalayas, southwestern China, and the northern part of Myanmar. They are known for their distinctive reddish-brown fur, white markings around their eyes, and bushy tails.}

        \vspace{2px}

        \textcolor{blue}{In the first picture, the red panda is curled up and appears to be sleeping or resting on a branch.} The background is blurred with green foliage, suggesting a natural, forested habitat. The focus is on the red panda's relaxed posture and the texture of its fur.

        \vspace{2px}
        
        \textcolor{blue}{In the second picture, the red panda is sitting upright on a branch, with its front paws resting on the branch. The animal is looking directly at the camera, giving a more alert and curious expression.} The background is also green and blurred, indicating a similar natural setting.

        \vspace{2px}
        
        \textcolor{blue}{The similarities between the two pictures include the presence of the red panda, the branch it is on, and the green, forested background. The differences lie in the red panda's posture and expression, with the first picture showing a more relaxed and the second a more alert state.} Additionally, the first picture has a more focused view of the red panda's fur and form, while the second picture captures the animal's face and eyes more clearly.
        
        \end{minipage}
        \end{tabular}
    \end{tcolorbox}
\end{minipage}
\caption{\textbf{Examples for Multi-Image Dialogue.} 
\textcolor{blue}{Blue} highlights the standout answers.
}
\label{fig:multi_image}
\end{figure*}

\noindent\textbf{Scientific Understanding}. Evaluating the capabilities of models in scientific understanding reasoning tasks is essential for advancing computational intelligence, particularly in contexts requiring in-domain knowledge and logical reasoning. In our study, we compared the performance of our InternVL 1.5 model with GPT-4V by administering complex multi-disciplinary problems designed to assess the accuracy of their reasoning. In Figure~\ref{fig:math_qa}, for the first question, both models accurately answered and provided an analysis from an aerodynamic perspective. For the second question, our model precisely analyzed the elements depicted in the image and provided the correct response, whereas GPT-4V speculated on the trend of amino acid transport. These results suggest that our method and GPT-4V exhibit comparable capabilities in scientific understanding and reasoning.

\noindent\textbf{Chinese Traditional Culture}. 
We selected two typical multimodal examples related to traditional Chinese art to evaluate our model.
As illustrated in Figure~\ref{fig:chinese}, both InternVL 1.5 and GPT-4V correctly recognize the Chinese traditional culture depicted in the image. Notably, InternVL 1.5 demonstrates a deeper understanding of this culture, as evidenced by its more detailed descriptions of the cultural elements in its response.

\noindent\textbf{Object Localization}.
Evaluating machine learning models for their proficiency in object localization tasks is essential, especially in applications requiring precise spatial awareness. In our comparative analysis, the performance of the InternVL 1.5 model was juxtaposed with GPT-4V, focusing on their ability to accurately detect and localize objects within various environments. Our assessments ranged from simple object recognition in cluttered scenes to complex scenarios involving dynamic interactions among multiple entities. As illustrated in Figure ~\ref{fig:localization}, the results demonstrate that InternVL 1.5 not only localized objects with high accuracy but also exhibited a comparable understanding of spatial relationships, matching the performance of GPT-4V.

\noindent\textbf{Multi-Image Dialogue}.
As shown in Figure~\ref{fig:multi_image}, in this experiment, we ask InternVL 1.5 and GPT-4V to compare the similarities and differences between the two images. 
As can be seen, both GPT-4V and InternVL 1.5 provide detailed and accurate responses.
Through this experiment, we discovered that although InternVL 1.5 was trained solely on single-image inputs, it exhibits strong zero-shot capabilities for multi-image dialogues.

\section{Conclusion}
\label{sec:conclusion}

This work introduced InternVL 1.5, an open-source MLLM designed to narrow the performance gap between open-source and proprietary models in multimodal understanding. By integrating a strong vision encoder with continuous learning capabilities, adopting a dynamic high-resolution strategy, and utilizing a high-quality bilingual dataset, InternVL 1.5 has demonstrated robust performance across a variety of benchmarks. Our evaluations indicate that the model achieves competitive performance with leading proprietary models, excelling particularly in OCR-related tasks and showing significant improvements in Chinese-related scene understanding.
While InternVL 1.5 has contributed to the open-source multimodal understanding, the field continues to evolve with many challenges ahead. We aspire to further enhance InternVL’s capabilities and invite collaboration with the global research community, hoping to enrich and expand the reach of open-source models together.

\clearpage
\newpage

\clearpage

{
    \small
    \bibliographystyle{ieeenat_fullname}
    \bibliography{main}
}
\end{CJK}
\end{document}